# A Survey of Source Code Representations for Machine Learning-Based Cybersecurity Tasks

BEATRICE CASEY, University of Notre Dame, Notre Dame, United States
JOANNA C. S. SANTOS, University of Notre Dame, Notre Dame, United States
GEORGE PERRY, University of Notre Dame, Notre Dame, United States

Machine learning techniques for cybersecurity-related software engineering tasks are becoming increasingly popular. The representation of source code is a key portion of the technique that can impact the way the model is able to learn the features of the source code. With an increasing number of these techniques being developed, it is valuable to see the current state of the field to better understand what exists and what is not there yet. This article presents a study of these existing machine learning based approaches and demonstrates what type of representations were used for different cybersecurity tasks and programming languages. Additionally, we study what types of models are used with different representations. We have found that graph-based representations are the most popular category of representation, and tokenizers and Abstract Syntax Trees (ASTs) are the two most popular representations overall (e.g., AST and tokenizers are the representations with the highest count of papers, whereas graph-based representations is the category with the highest count of papers). We also found that the most popular cybersecurity task is vulnerability detection, and the language that is covered by the most techniques is C. Finally, we found that sequence-based models are the most popular category of models, and Support Vector Machines are the most popular model overall.

## 1 Introduction

Software vulnerabilities are defects that affect a software system's intended security properties [1] which allow attackers to perform malicious actions. As our lives become more attached to technology, software vendors are increasingly pressured into engineering secure software systems—that is, taking proactive measures of preventing/repairing vulnerabilities prior to deploying the systems into production. There are several practices to address security concerns before software release in each phase of the software development lifecycle. In the requirements phase, *misuse/abuse cases* [2, 3] and *threat modeling* [4] are useful to better understand the security requirements by

Authors' Contact Information: Beatrice Casey, University of Notre Dame, Notre Dame, Indiana, United States; e-mail: bcasey6@nd.edu; Joanna C. S. Santos, University of Notre Dame, Notre Dame, Indiana, United States; e-mail: joannacss@nd.edu; George Perry, University of Notre Dame, Notre Dame, Indiana, United States; e-mail: gperry@nd.edu.

identifying the potential threats to the system and possible ways to mitigate them. During the design phase, *architectural risk analysis* [5] can be used to assess the likelihood of exploitation of an asset and *security tactics* and *patterns* [6] can be applied in the design of the software as a proven solution that works under a context. At the implementation phase, *secure code reviews* can be performed to systematically examine the source code with the goal of identifying and fixing vulnerabilities [7]. In the testing phase, *penetration testing* aims to exercise the software in many ways in an attempt to break it and discover vulnerabilities [8] and *static/dynamic analysis* tools can be used to identify potential vulnerabilities in the source code [9, 10].

Although these practices can help improve a software system's security, it can be error prone and time consuming for engineers to perform them. For example, finding vulnerabilities in code can be difficult for engineers, especially when they do not know what to look for [7]. With the advances of **machine learning (ML)**, prior works have applied ML techniques to several of these cybersecurity tasks, such as vulnerability detection [11–16], malware detection [17–26] and malicious behavior detection [27]. These techniques are valuable because they can help improve the security of code that is released and speed up otherwise error-prone and time-consuming tasks. For example, a model that detects vulnerabilities prior to deployment would save time and money, and increase the security of the system as a whole, particularly since developers are oftentimes unaware when a vulnerability exists in code until it is exploited or found by security analysts [28].

*ML models* are unable to understand raw source code—that is, source code in an unprocessed form. As ML models rely on numerical data to adjust the weights necessary for learning, source code must be represented in a form that effectively captures both its *structural* and *semantic* information. This transformation allows the model to understand and learn from the underlying patterns in the code. Thus, *source code representation* is a crucial part during the development of ML-based techniques because different representations will offer different information that the model learns from to perform their task [29].

There are many ways in which source code can be represented—for example, as an **Abstract Syntax Tree (AST)** [12, 16, 30–34], **Control Flow Graph (CFG)** [12, 13, 35–37], or tokenized [11, 15, 38–40], among others. Although prior works [41–48] have introduced novel source code representations for different cybersecurity tasks, there is no current understanding of what representations exist and are commonly used, as well as the cybersecurity tasks the model is being used for. Furthermore, many of these ML-based techniques are only tested on or made for a particular programming language, and there is no current understanding of the languages that are covered by these techniques. Understanding the available source code representations and what they offer will allow researchers to identify which representation they may want to use based on what task they aim to complete. Additionally, understanding the relationship between cybersecurity tasks and representations will allow researchers to either choose a representation that has previously been used for a particular task or test if a different representation would be more suitable.

In this article, we conduct a **Systematic Literature Review (SLR)** by following the guidelines by Kitchenham and Charters [49] to understand the current state of the art of source code representation for ML-based techniques for cybersecurity tasks. We investigate the popularity of certain representations and cybersecurity tasks, the programming languages covered by existing techniques, and the common types of ML models used with different representations. We also investigate relationships between representations and cybersecurity tasks. The goal of the study is to allow researchers to understand the gaps in this domain, particularly if there are certain cybersecurity tasks, or languages that are neglected by existing techniques. Additionally, we study and contrast existing representations.

The contributions of this article are as follows: (1) an examination of the state of the art of ML-based cybersecurity tasks; (2) an investigation of source code representations, and their relationships to cybersecurity tasks and models; (3) the identification of the programming languages that are covered/not covered by existing ML-based techniques; and (4) a comparison of the different source code representations. The article's artifacts, including datasets, code, and additional resources, are available on GitHub: https://github.com/s2e-lab/code-representation-slr.

The rest of this article is organized as follows. Section 2 provides the definitions of terms that are relevant for understanding our work. Section 3 describes related work. Section 4 explains the methodology of this SLR. Sections 5 through 9 share the results of this work. Section 10 explains threats to validity, and Section 11 provides a discussion and conclusion of our findings.

## 2 Background

This section discusses core terminology such that the article can be understood by a broader audience.

### 2.1 ML for Secure Software Engineering

Along with developing new code of good quality, software engineers are responsible for discovering and resolving bugs, defects, vulnerabilities, and any other issues that could arise in source code. These tasks can be difficult, error prone, time consuming, and tedious [7]. Following the definition of Kemmerer [50], a *cybersecurity task* is a task that is aimed at thwarting would-be intruders. Thus, the cybersecurity tasks that software engineers work on today involve resolving and identifying code that could allow an attacker to take advantage of a system.

With the recent advances of ML, several prior works identified ways in which ML could assist software engineers in these cybersecurity tasks [51]. In particular, given how negatively vulnerabilities and other security-related issues impact companies, researchers started looking into how ML can help mitigate these issues and overall improve the quality of source code that is put out to the public.

### 2.2 Source Code Representations and Code Embeddings

A *source code representation* captures the source code's *syntax* and *semantics* such that the model is able to learn the key features. There are many ways a piece of source code can be represented. For example, given the structured nature of source code, prior works captured this structure by representing it as a *graph* [52]. Other works [38, 53, 54] used **Natural Language Processing (NLP)** techniques on source code to leverage technology and knowledge that already exists. These representations offer different information about the source code and thus impact what the model can learn about it. For example, NLP techniques do not offer structural information, but they provide semantic information. Therefore, the model will learn the semantic relationships, but not the structural relationships in the code.

ML models learn from *vector embeddings*, which is a low-dimensional way to represent high-dimensional data. In the case of learning source code, the embeddings are created from the source code representation. The source code representation is the *first* level of abstraction for the original source code. This is what is considered as the *feature extraction* phase. In this phase, the original high-dimensional data are transformed into lower-dimensional data, which represents the key features of the data. The feature extraction is aimed at preserving as much of the information about the original data as possible [55]. The vector embeddings are the *second* level of abstraction and are what allows us to perform ML techniques on the representations by transforming the representation to the numerical form that machines are able to understand [56]. In this literature review, we focus only on the first level of abstraction (i.e., source code representation).

There are a multitude of techniques to create these embeddings. From the realm of NLP work, the embeddings are usually created by using the WORD2VEC model [57], which takes tokens and converts them to numerical vectors. Inspired by this methodology, researchers have found a way to take a graph and create these vector embeddings (GRAPH2VEC [58]). This technique is what is typically used to generate embeddings from the tree or graph structures generated by these source code representations. GRAPH2VEC implements ideas from DOC2VEC and WORD2VEC, and treats a whole graph as a document and the subgraphs as words [58]. Additionally, other works have looked at how to create embeddings straight from source code and created techniques such as CODE2VEC [59] and GRAPHCODE2VEC [60].

## 3 Related Work

To the best of our knowledge, this is the first-of-its-kind SLR that focuses on the representations of source code used in ML-based techniques for cybersecurity tasks. There are a number of papers [61–69] that perform either a systematic mapping study or a survey of the literature in ML for software engineering, with some papers focusing on vulnerability detection, analysis, or assessment. All of these papers primarily focus on the *ML models* used, but few mention or give detailed descriptions of the representations used in these efforts. Additionally, some of the papers focus only on deep learning techniques [64–66]. Unlike these prior works, our survey article has its primary focus on the topic of representations used for ML for security-related tasks.

Ghaffarian and Shahriari [70] surveyed techniques used for vulnerability analysis and outlined four main categories that the approaches of these techniques fall under: *software metrics*, *anomaly detection*, *vulnerable code pattern recognition*, and *miscellaneous*. Similarly, Nazim et al. [63] analyzed deep learning models for vulnerable code detection. Their study examined the different dataset types used to train deep learning models (synthetic, semi-synthetic, real data, etc.), the evaluation metrics used to assess performance, as well as the different source code representations used. Unlike these prior works, we look beyond vulnerability analysis and detection, and instead look at *all* security-related software engineering tasks. We also have a broader scope because we look at all types of ML models, not just deep learning models.

Wu [71] performed a literature review on NLP techniques for vulnerability detection. While this paper does give a brief overview of different types of representations, its focus is on NLP techniques, particularly NLP models that are focused for code intelligence, such as CodeBERT and CodeXGlue, in the instance of vulnerability detection. Chen and Monperrus [72] perform a similar literature review, investigating word embedding techniques on programs. In this survey, the authors explore different granularities of embeddings from different papers and show visualizations of the different embeddings. We focus on a wide array of security-related tasks, as well as many representations and how they can impact the model's ability to learn vulnerabilities.

Kotti et al. [73] performed a tertiary study on ML for software engineering. This paper evaluated 83 reviews, or surveys, on the field of ML for software engineering. While our article focuses on *security-related* software engineering tasks, Kotti et al. [73] investigated all ML-based software engineering tasks. Additionally, while this paper is a broader analysis of the software engineering tasks that ML covers, our article focuses primarily on the representations of source code used to perform a security task.

Two papers create a taxonomy for software engineering tasks and ML, with one paper focusing broadly on software engineering challenges for ML systems [74], and the other focusing on software vulnerability detection and ML approaches [75]. Neither of these papers delve into source code representation efforts and what information they offer. However, Hanif et al. [75] do mention the importance of representation in the ML pipeline. Unlike these studies, our article focuses on the representations used, rather than just the models, and focuses on a wide array of security tasks.

Usman et al. [76] performed a survey on representation learning efforts in cybersecurity. However, this paper does not focus on the representation of source code, but rather different ML algorithms used for cybersecurity issues, as well as datasets and how industry utilizes these different efforts to improve their cybersecurity. Similarly, Macas et al. [77] created a survey on deep learning techniques for cybersecurity. This article focuses on deep learning techniques to analyze internet traffic. It provides insights and future directions in this area of deep learning to analyze internet traffic for cybersecurity.

## 4 Methodology

We followed the guidelines outlined by Kitchenham and Charters [49] to conduct our SLR, which involves three major activities: *planning*, *conducting*, and *reporting* the review. During the *planning* phase, we defined this study's research questions and the search query used to find papers. During the *conducting* phase, we searched three library sources and downloaded all the papers we found into CSV files. We then applied our *inclusion* and *exclusion* criteria in three phases to eliminate papers until we got to the final group of papers that are included in this study. Two reviewers independently read each paper and performed an analysis, extracting the information that is relevant to the research questions we developed in the planning phase. We reviewed and resolved discrepancies to get the final analyses. We calculated Cohen's kappa to evaluate the reliability of our evaluation. Our score of 0.97 indicates that we had a near-perfect agreement in our analysis. Finally, during the *reporting* phase, we analyzed our data and organized it so that we could answer the research questions we posed.

### 4.1 Research Questions

Through this SLR, we aim to answer five research questions.

**RQ1:** *What are the most commonly used source code representations?*

In this first question, we investigate the source code representations that were used for solving security problems. We aim to understand which source code representations are more popular and compare their tradeoffs.

**RQ2:** *Do certain cybersecurity tasks only or mostly use one type of source code representation?*

In this question, we investigate what source code representations are being used for each cybersecurity task and whether certain representations are favored for specific tasks. Along with this, we want to investigate why it would be the case that a particular representation is preferred for a task.

**RQ3:** *What cybersecurity tasks are covered by the techniques that have been created?*

We investigate how these tasks fit into the software development lifecycle to identify how these techniques would be used during software development. Furthermore, we describe and find every cybersecurity-related task so that we provide a clearer picture of the task and its importance in the realm of software security.

**RQ4:** *What programming languages are predominantly targeted by the ML-based techniques for cybersecurity tasks?*

Given the vast number of programming languages used in practice, we investigate what languages the source code analyzed by these techniques is written. This question helps reveal any gaps in the coverage of programming languages by these techniques.

**RQ5:** *What models are commonly used with different representations?*

In this last question, we study which source code representations are used for different model types. By examining the frequency co-occurrence between models and representations, we aim to understand common trends and preferences in the design of ML-based techniques. This analysis helps identify potential gaps and opportunities for improvement in how models are paired with specific representations.

### 4.2 Search Method

We used the following search string to find all *primary* studies related to the representation(s) of source code for ML-based cybersecurity tasks: `(“machine learning”` **`OR`** `“deep learning”` **`OR`** `“artificial intelligence”)` **`AND`** `(“security”` **`OR`** `“vulnerability”)` **`AND`** `(“code”)`. While this is a very general search string, which resulted in a total of 67,512 papers, we decided that rather than having a specific string that may miss a category of software security tasks or representations, we would make a general string and manually eliminate any papers that do not meet our inclusion criteria, or fit our exclusion criteria.

We searched three databases to find relevant papers: the ACM Digital Library,[1] IEEE Xplore,[2] and Springer Link.[3] We also searched nine ML and NLP conferences which are considered to be A* conferences by the CORE conference ranker. These nine conferences are National Conference of the American Association for Artificial Intelligence (AAAI), Conference on Learning Theory (COLT), International Conference on Learning Representations (ICLR), International Conference on Machine Learning (ICML), International Joint Conference on Artificial Intelligence (IJCAI), Advances in Neural Information Processing Systems (NeurIPS, formerly known as NIPS), Association for Computational Linguistics (ACL),[4] Empirical Methods in Natural Language Processing (EMNLP), and International Conference on the Principles of Knowledge Representation and Reasoning (KR). We did not search papers published at the International Joint Conference on Autonomous Agents and Multiagent Systems (previously the International Conference on Multiagent Systems, ICMAS, changed in 2000), the ACM International Conference on Knowledge Discovery and Data Mining, or the IEEE International Conference on Data Mining because these conferences' proceedings were included in the ACM and IEEE libraries.

### 4.3 Inclusion and Exclusion Criteria

Table 1 lists the inclusion/exclusion criteria applied to the papers in multiple stages to eliminate papers irrelevant for this study. We limited our search to papers published between January 2012 and May 2023. Our inclusion criteria focused on ML-based techniques for cybersecurity that involved representing source code. We eliminated duplicate studies, works not in English, and any papers that were not full papers, such as books, short papers (i.e., papers with less than five full pages of text, not including references), and tutorials. We also disregarded any papers that did not represent the source code itself (e.g., papers that dealt with binary files, extracting from the Android Manifest), as we are interested in only understanding the representation of raw source code.

[1] https://dl.acm.org
[2] https://ieeexplore.ieee.org
[3] https://link.springer.com
[4] When we got the data for ACL, that includes *Transactions of the Association for Computational Linguistics* and ACL Anthology.

Table 1. Inclusion and Exclusion Criteria

| Inclusion Criteria | Exclusion Criteria |
|---|---|
| **I1** Written between 2012 and May 2023<br>**I2** A full paper<br>**I3** Focused on ML for cybersecurity tasks<br>**I4** Contains information regarding the source code's representation | **E1** Duplicated studies<br>**E2** Books, reference work entries, reference works<br>**E3** Position papers, short papers, tool demo papers, keynotes, reviews, tutorials, and panel discussions<br>**E4** Studies not in English<br>**E5** Survey/comparative studies. |

Fig. 1. Overview of the Three Stages of our Search Process

### 4.4 Paper Selection

Figure 1 shows the number of papers that made it through each stage of the selection process. We started out with 67,512 total primary studies. We first excluded duplicate studies and studies that were outside the year range of 2012 to May 2023 (criterion I1), as well as non-full papers (criterion E1). This resulted in 52,836 papers. Subsequently, we inspected each paper's title, keywords, and abstract to include/exclude papers based on whether they fit our criteria. After this search, we were left with 520 papers. We then applied the same criteria on these 520 papers, this time by reading the full paper. This left us with the 141 papers that are included in this survey.

### 4.5 Data Extraction

As we went through the papers, we extracted the key information we were looking for to answer our research questions: the *representation* used, the *cybersecurity task* it was completing, the *programming languages* the technique was designed for or tested on, and the *model type* used in the work. When determining the representation used, we used a *one-paper-to-many-representations* approach based on what the paper claims to be the representation(s) used. In other words, we consider every representation that a paper explicitly uses to train/fine-tune their model(s) as a separate, valid representation. However, in the case a paper employs one or more representations solely as internal steps to derive a final representation, then we treat only the final representation as the representation used. For example, if a paper refines an initial CFG through multiple transformations to produce a regular expression to represent the code, we capture only *regular expression* as the representation being used by the paper.

Besides capturing the preceding metadata to directly answer our five research questions, we conducted a comprehensive analysis of the performance of the source code representation(s). Specifically, we went through the reported results in each paper, extracting the metrics provided. This allowed us to aggregate the data and compare the performance of different source code representations across various cybersecurity tasks.

## 5 RQ1 Results: What Are the Commonly Used Source Code Representations?

Table 2 summarizes the source code representations used/described by the surveyed papers. Similar to a prior work [64], we organized these representations into four categories: *graph based*, *tree based*, *lexical*, and *miscellaneous*. We found that the three most commonly used source code representations are a *tokenizer*, an *AST*, and *code metrics*. We also found two papers [51, 178] that used a slightly modified AST version to represent source code (we denoted them as AST+ in Table 2). Although more rare, some papers [21, 87, 92] represented code as an *image* analysis, and performed an image analysis of source code to identify patterns that are associated with vulnerabilities or malware. The following subsections give a detailed explanation of what each representation is and the information it carries out to the embeddings.

Table 2. Source Code Representations, Their Categories, and Frequency of Use in the Surveyed Papers

|  | Representation | Papers | # |  | Representation | Papers | # |
|---|---|---|---|---|---|---|---|
| Graph Based | Control Flow Graph (CFG) | [12, 13, 35–37, 78–86] | 14 | Graph Based | Component Behavior Graph (CBG) | [27] | 1 |
| Graph Based | Program Dependence Graph (PDG) | [12, 87–93] | 8 | Graph Based | Component Dependency Graph (CDG) | [27] | 1 |
| Graph Based | Data Flow Graph (DFG) | [13, 37, 78, 80, 84, 94, 95] | 7 | Graph Based | Contextual ICFG (CICFG) | [26] | 1 |
| Graph Based | Call Graph | [20, 22, 25, 35, 79, 96] | 6 | Graph Based | Contextual Permission Dependency Graph (CPDG) | [26] | 1 |
| Graph Based | Code Property Graph (CPG) | [12, 97–100] | 5 | Graph Based | Contextual Source and Sink Dependency Graph (CSSDG) | [26] | 1 |
| Graph Based | Interprocedural Control Flow Graph (ICFG) | [18, 19, 26] | 3 | Graph Based | Crucial Data Flow Graph (CDFG) | [101] | 1 |
| Graph Based | Contextual API Dependency Graph (CADG) | [19, 26] | 2 | Graph Based | Program Graph | [102] | 1 |
| Graph Based | Contract Graph | [46, 103] | 2 | Graph Based | Propagation Chain | [104] | 1 |
| Graph Based | Program Slices | [14, 105] | 2 | Graph Based | Property Graph | [45] | 1 |
| Graph Based | Simplified Code Property Graph (SCPG) | [106, 107] | 2 | Graph Based | Semantic Graph | [108] | 1 |
| Graph Based | System Dependence Graph (SDG) | [109, 110] | 2 | Graph Based | Slice Property Graph (SPG) | [111] | 1 |
| Graph Based | Token Graph | [112, 113] | 2 | Graph Based | Value Flow Graph (VFG) | [44] | 1 |
| Graph Based | Code Aggregate Graph (CAG) | [114] | 1 | | | | |
| Lexical | Tokenizer | [11, 15, 17, 38–40, 43, 53, 54, 82, 86–90, 104, 115–145] | 47 | Miscellaneous | Code Metrics | [96, 116, 132, 133, 146–157] | 16 |
| Lexical | intermediate code and Semantics-based Vulnerability Candidate (iSeVC) | [48, 158, 159] | 3 | Miscellaneous | Code Gadgets | [14, 41, 110, 160–163] | 7 |
| Lexical | source code and Syntax-based Vulnerability Candidate (sSyVC) | [48, 158] | 2 | Miscellaneous | Image | [21, 87, 92] | 3 |
| Lexical | Contract Snippet | [47] | 1 | Miscellaneous | Opcode Sequences | [164] | 1 |
| Tree Based | Abstract Syntax Tree (AST) | [12, 16, 30–34, 37, 78, 80, 85, 88–90, 106, 142, 144, 157, 163, 165–177] | 32 | Miscellaneous | Application Information | [23] | 1 |
| Tree Based | AST+ | [51, 178] | 2 | Miscellaneous | API Calls | [24] | 1 |
| Tree Based | Parse Tree | [179] | 1 | Miscellaneous | Regular Expression | [180] | 1 |

## 5.1 Tree-Based Source Code Representations

*Tree-based* representations are those that demonstrate the *hierarchical* nature of source code [64].

*5.1.1 Abstract Syntax Tree.* An *AST* is a tree representation of source code that provides information about code elements (e.g., variables) and their structural relationship [63, 181]. It was the most popular representation (used by 32 papers). Although 1 paper [168] used code2vec [182] as a way to generate source code embeddings, the basis of their model is an AST; the source code is represented as an AST before the vectors are generated.

Using the information from an AST, models can capture general structural code patterns, since ASTs abstract away the low-level syntax details of the underlying programming language of the code [12]. This reduces learning effort and allows for ASTs to be used for multiple tasks [59]. Embeddings for ASTs can be generated in different ways. Typically, the node and path are what form the embedding so that the relationship between two nodes can be effectively captured by the embedding [59].

*5.1.2 Parse Tree.* A *parse tree* represents the hierarchy of *tokens*—that is, the program's terminal and non-terminal symbols. This data structure is generated by the language's parser [183]. Thus, the nodes represent the derivation of the grammar that yields the input strings. This representation has been used by Ceccato et al. [179] to represent SQL queries to train a model that detects SQL injection vulnerabilities. Although parse trees and ASTs both represent source code in a tree structure, their key difference is that ASTs are much *simpler* than parse trees, as they abstract away grammar-related nodes while parse trees retain these tokens and their meanings with respect to their grammar. In Appendix A.1, we provide an example of a parse tree and AST for the same source code to demonstrate these differences.

*5.1.3 AST+.* One paper [178] used a representation that is an enhanced version of an AST (which we denote in our article as AST+). That work uses a convention [184] that describes AST nodes in three types: *placeholder*, *API*, and *syntax nodes*. The ASTs are serialized and traversed using depth first traversal, and each node and element is mapped to a vector. Xia et al. [51] do not specify the modifications made to the AST; however, the paper states that additional edges are added to the AST to capture more semantic and stream information. This representation was used for vulnerability detection [51, 178].

## 5.2 Graph-Based Representations

*Graph-based* representations are those that transform source code into a *graph* form, with nodes and edges representing certain characteristics and relationships, respectively, between each code element. Graph-based representations can be embedded using Graph2vec [58], as it is an optimized method to transform the graphs into the low-dimensional numerical vectors that the models will learn from. As shown in Table 2, we found 24 different graph-based representations, whose descriptions are provided in the next sections.

*5.2.1 Control Flow Graph.* A *CFG* [185] was the most popular graph representation used in 14 papers [12, 13, 35–37, 78–86]. A CFG is a directed graph $g = (V, E)$ with nodes $v \in V$ and edges $e \in E$, where $E \subseteq V \times V$. The set of nodes $V$ represents the *basic blocks* of a program's procedure (i.e., a function/method), whereas the edge set $E$ represents the control flow between the basic blocks. A basic block is a group of instructions that are executed in order, one after the other. A CFG's edge $e = v_{src} \longrightarrow v_{dst}$ denotes that the program's execution can flow from $v_{src}$ to $v_{dst}$.

*5.2.2 Interprocedural Control Flow Graph.* An **Interprocedural Control Flow Graph (ICFG)** is a variation of the CFG that describes not only *intraprocedural flows* among the basic blocks but also *interprocedural* ones [19]. An ICFG connects individual CFGs at the call sites to represent control flows across procedures. Thus, the ICFG allows the model to understand the control flow of the *whole program*, whereas the CFG allows the model to understand the control flow of a specific *procedure* [186].

*5.2.3 Data Flow Graph.* Seven papers used a **Data Flow Graph (DFG)**, which is a graph $G = (V, E)$, where the nodes $v \in V$ are statements in the source code, and the edge set $e \in E$ represents the data dependencies between the nodes. In other words, an edge $e = v_{src} \longrightarrow v_{dst}$ indicates that the node $v_{dst}$ uses data that has been defined by $v_{src}$.

*5.2.4 Program Dependence Graph.* **Program Dependence Graphs (PDGs)** were the second most popular graph-based representation, being used by eight papers. A PDG [187] is a directed graph $g = (V, E)$ that shows the *data* and *control* dependencies for each statement in a program's procedure. The set of nodes $V$ in a PDG is partitioned into two types: *statement* nodes $V_{stmt}$ and *predicate expression* nodes $V_{pred}$. A statement node $v_{stmt} \in V_{stmt}$ represents simple statements in a program that are actions to be carried out by a program (e.g., `x = 2;`). A predicate node $v_{pred} \in V_{pred}$ denotes statements that evaluate to *true* or *false* (e.g., `x != 2`). The edge set $E$ in a PDG has two partitions: *control dependency* edges $E_c$ and *data dependency* edges $E_d$. A control dependency edge $e_c = v_{src} \longrightarrow v_{dst}$ indicates that $v_{dst}$ only executes if the predicate expression in $v_{src}$ evaluates to true. A data dependency edge $e_d = v_{src} \longrightarrow v_{dst}$ denotes that $v_{dst}$ uses data that has been defined by $v_{src}$.

*5.2.5 Call Graph.* A *call graph* is a directed graph $g = (V, E)$ in which the nodes are the functions/methods in a program, whereas the edges represents caller-callee relationships among program's procedures [188]. An edge $e = v_{src} \longrightarrow v_{dst}$ denotes that $v_{src}$ invokes $v_{dst}$. These graphs can be of two types: *static* and *dynamic* call graphs. Dynamic call graphs give information regarding the procedure calls of a program while it is being executed. It shows the sequence of function/method calls, and the parameters that are passed to each procedure in the sequence. Static call graphs only give information about the *potential* execution paths a program can have based on information available at compile time. Thus, a static call graph is not as accurate in reflecting the actual calls in a program, particularly if the program is complex [188]. A total of six papers included in this survey used static call graphs [20, 22, 25, 35, 79, 96].

In Appendix A.2, we provide an example of a CFG, ICFG, DFG, PDG, and call graph to show the different type of information each representation provides for the same code snippet.

*5.2.6 System Dependency Graph.* Two papers represented source code using **System Dependency Graphs (SDGs)** [109, 110]. An SDG is a graph with multiple PDGs connected via the caller-callee relation given by a call graph. SDGs extend PDGs by describing the *interprocedural* relationships between the program's *entrypoints*[5] and the procedures they call [109]. To connect the PDGs, there are additional nodes and edges which dictate the *actual input parameters* and *actual output values* of a procedure. Every passed argument has an *actual-in* node $a_i$ and a *formal-in* node $f_i$ which are connected by the parameter-in edge $a_i \longrightarrow f_i$. Every modified parameter and returned value has an actual-out node $a_o$ and a formal-out node $f_o$, which are connected by the parameter-out edge $f_o \longrightarrow a_o$. The formal-in and -out nodes are control dependent on the entry node $e$ and actual-in and -out nodes are control dependent on the call node $c$. This parameter passing model ensures that interprocedural events of a procedure are propagated by the call sites.

*5.2.7 Program Slices.* A *program slice* [189] is a subgraph of a PDG or SDG that includes only the nodes that are relevant to a computation at a specific point in the program. This subgraph is computed using a *slicing criterion* $\langle v, p \rangle$ which denotes a variable of interest $v$ at a program point $p$. These slices can be computed in a *backward* or *forward* fashion. A *backward slice* includes all the nodes that may affect the value of $v$ at the program point $p$. A *forward slice* includes all nodes that are affected by the variable $v$ at the program point $p$. Program slices were used by two papers [14, 105] to detect vulnerabilities. The slicing criterion is determined by statements in the code that are considered as vulnerable. The statements could also be points where values are changed, which could then lead to an API call being vulnerable. Cheng et al. [105] use a PDG and perform forward and backward slicing from the node of interest. It is not specified in the work of Chen et al. [14] whether a PDG or SDG is used, but the same criterion for backward and forward slicing are used (i.e., forward slices include statements that are affected by the node of interest, and backward slices include statements that affect the node of interest).

*5.2.8 Crucial Data Flow Graph.* A **Crucial Data Flow Graph (CDFG)**, introduced by Wu et al. [101], is a subgraph of a DFG that contains only the *crucial* information from the DFG that could trigger a reentrancy vulnerability in smart contracts. The *crucial* nodes are variables containing sensitive or critical information, and that have a direct data flow to another crucial node. A CDFG is defined as $CDFG = (V, E)$, where $v \in V$ are the crucial nodes and the edges $e \in E$ represent the data flow relationship. For example, $e = v_{src} \rightarrow v_{dst}$ indicates that $v_{src}$ and $v_{dst}$ are both crucial nodes, and that there is a data flow between the two variables.

[5]Entrypoints are the functions/methods that starts the program execution (e.g., the `main()` function in C).

In Appendix A.2, we provide an example of an SDG, a program slice, and a CDFG as a demonstration for what information can be expected from each representation.

*5.2.9 Program Graph.* Wang et al. [102] introduced the concept of the *program graph*, which is a directed graph $g = (V, E)$ in which the nodes $v \in V$ can be *statements*, *identifiers* (e.g., function declarations or variables), or *values*. This graph has eight types of edges: *control flow* edges, *data flow* edges, *guarded by* edges, *jump* edges, *ComputedFrom* edges, *NextToken* edges, *LastUse* edges, and *LastLexicalUse* edges. A control flow edge $e_{ctr} = v_{src} \rightarrow v_{dst}$ indicates that $v_{dst}$ can execute after $v_{src}$. A data flow edge $e_{data} = v_{src} \rightarrow v_{dst}$ indicates that $v_{dst}$ uses a variable that has been defined by $v_{src}$. A guarded by edge $e_g = v_{src} \rightarrow v_{dst}$ indicates that $v_{dst}$ only executes if the expression in $v_{src}$ evaluates to true (which is useful to identify operations that may be in the wrong order). A jump edge $e_j = v_{src} \rightarrow v_{dst}$ indicates that $v_{dst}$ has a control dependency from $v_{src}$. A ComputedFrom edge $e_{compFrom} = v_{src} \rightarrow v_{dst}$ indicates that $v_{src}$ is or contains a variable used in an expression in $v_{dst}$. A NextToken edge $e_{next} = v_{src} \rightarrow v_{dst}$ indicates that $v_{dst}$ is a successor of (i.e., follows) $v_{src}$, where $v_{dst}$ and $v_{src}$ are terminal nodes or tokens from the AST. A LastUse edge $e_{last} = v_{src} \rightarrow v_{dst}$ indicates that $v_{dst}$ uses the same variable that is used in $v_{src}$. A LastLexicalUse edge $e_{lastLex} = v_{src} \rightarrow v_{dst}$ indicates that $v_{dst}$ uses the same variable that is used in $v_{src}$ if $v_{src}$ is an `if` statement.

*5.2.10 Code Property Graph.* Used by five papers [12, 97–100], a **Code Property Graph (CPG)** is a combination of ASTs, CFGs, and PDGs [52]. It was first introduced by Yamaguchi et al. [52] specifically as a way to detect vulnerabilities in C/C++ programs using static analysis. The way a CPG is generated is by taking the AST, CFG, and PDG of a program, modeling them as *property graphs*, and then these models are jointly combined by connecting statement and predicate nodes. A CPG is formally defined as $g = (V, E, \lambda, \mu)$, which is a directed, labeled, attributed multigraph, with nodes $v \in V$, edges $e \in E$, edge labeling function $\lambda$, and a property mapping function $\mu$. The set of nodes $V$ in a CPG is the nodes from an AST. The edge set $E$ in a CPG has three partitions: *control flow edges* $E_{cf} \subseteq V \times V$, *program dependency* edges $E_{pd} \subseteq V \times V$, and *AST* edges $E_{ast} \subseteq V \times V$. A control flow dependency $e_{cf} = v_{src} \rightarrow v_{dst}$ indicates that $v_{src}$ can flow to $v_{dst}$ in the next step of the program. A program dependency edge $e_{pd} = v_{src} \rightarrow v_{dst}$ indicates that $v_{dst}$ has a program dependence edge from $v_{src}$. An AST edge $e_{ast} = v_{src} \rightarrow v_{dst}$ indicates that $v_{dst}$ is syntactically related to $v_{src}$. The *edge labeling function* $\lambda : E \rightarrow \Sigma$ assigns a label from the alphabet $\Sigma$ to each edge in $E$. The function $\mu : (V \cup E) \times K \rightarrow S$ applies properties to nodes and edges, where $K$ is the set of property keys and $S$ is the set of property values. Since a CPG is a combination of so many representations, it provides a very robust understanding of code. Other implementations of a CPG enhance it by adding information from a DFG [63].

*5.2.11 Simplified Code Property Graph.* While CPGs are able to capture rich semantic and syntactic information, it is also very complex to create. Generating a PDG alone has a complexity of $O(n^2)$. The size of the graphs are also rather large, one example having 52 million nodes and 87 million edges [52]. To solve this issue, two papers implemented a **Simplified Code Property Graph (SCPG)** [106, 107]. An SCPG only uses edges from an AST and a CFG, as data dependence can be approximated from these two graphs. The nodes in the SCPG have two values: the *code tokens* and the *node type*. Removing the need to generate a PDG greatly reduces the cost of generating this representation, as one would only need to generate the AST and CFG.

*5.2.12 Property Graph.* A *property graph* [45] is variant of a CPG, defined as $g = (V, E, \Lambda, \Sigma, \mu, \lambda, \sigma)$. Here, the edges and nodes are the same as the CPG. The adjacency function $\mu : E \rightarrow V \times V$ maps any edge to an ordered pair of its source and destination vertices. The function $\lambda : V \longrightarrow \Lambda$ maps any given vertex to its respective attributes $\Lambda$, and $\sigma : E \longrightarrow \Sigma$ is the attribute function for edges, which, just as $\lambda$, maps any given edges to its respective attributes $\Sigma$.

*5.2.13 Code Aggregate Graph.* A **Code Aggregate Graph (CAG)** is built from a combination of an AST, CFG, PDG, dominator tree, and post-dominator tree. A CAG is formally defined as $g = (V, E)$, which is a directed labeled, attributed multigraph, with nodes $v \in V$ and edges $e \in E$ where $E \subseteq V \times V$. The set of nodes $V$ in a CAG is the nodes from an AST. The edge set $E$ in a CAG has five partitions: *control flow* edges $E_{cf} \subseteq V \times V$, *program dependency* edges $E_{pd} \subseteq V \times V$, *AST* edges $E_{ast} \subseteq V \times V$, *dominator tree* edges $E_{dt} \subseteq V \times V$, and *post-dominator tree* edges $E_{pdt} \subseteq V \times V$. A control flow dependency $e_{cf} = v_{src} \longrightarrow v_{dst}$ indicates that $v_{src}$ can flow to $v_{dst}$ in the next step of the program. A program dependency edge $e_{pd} = v_{src} \longrightarrow v_{dst}$ indicates that $v_{dst}$ has a program dependence edge from $v_{src}$. An AST edge $e_{ast} = v_{src} \longrightarrow v_{dst}$ indicates that $v_{dst}$ is syntactically related to $v_{src}$. A dominator tree edge $e_{dt} = v_{src} \longrightarrow v_{dst}$ indicates that the operation $v_{dst}$ is dominated by $v_{src}$ and all of $v_{src}$ dominators (i.e., all paths from the entry node to $v_{dst}$ first pass through $v_{src}$). A post-dominator tree edge $e_{pdt} = v_{src} \longrightarrow v_{dst}$ indicates that the operation $v_{dst}$ is post dominated by $v_{src}$, meaning that all paths from $v_{dst}$ to the end node must pass through $v_{src}$. Using a dominator tree and post-dominator tree allows this representation to better capture semantic information in source code. This, in turn, allows models to perform better in the task of vulnerability detection. Nguyen et al. [114] point out certain information that a CFG and an AST in particular fail to capture, and how a dominator tree and a post-dominator tree can better describe these attributes.

*5.2.14 Value Flow Graph.* A **Value Flow Graph (VFG)** is similar to a PDG in that is shows the interprocedural program dependence. The edges, just like in a PDG, describe the control flow and data dependency of the program [190]. A VFG $g = (V, E)$ is a directed labeled graph, with nodes $v \in V$ and edges $e \in E$. The set of nodes $V$ is a pair $(\gamma_1, \gamma_2)$ in which $\gamma_1$ is a node from the pre-directed acyclic graph and $\gamma_2$ is a node from the post-directed acyclic graph. Both $\gamma_1$ and $\gamma_2$ represent the same value. The edge set $E$ in a VFG is a pair $(v, v')$ such that $N(v)$, the node flow graph of $v$, is the predecessor of $N(v')$, the node flow graph of $v'$, and values are maintained along the connecting edge [191].

The paper [44] that used a VFG uses a special process that selects and preserves feasible value-flow paths to reduce the amount of data needed for training models for path-based vulnerability detection. This makes their method more lightweight than a typical VFG would be.

In Appendix A.2, we provide an example of a CPG, a CAG, and a VFG for the same source code to highlight the differences in the type of information provided by each representation.

*5.2.15 Component Dependency Graph.* A **Component Dependency Graph (CDG)** [27] represents the relationships between the different components in a graph and was created to capture Android app program logic. The CDG is formally defined as $g = (V, E)$, which is a directed labeled graph with nodes $v \in V$ and edges $e \in E$. The set of nodes $V$ in a CDG represents the components of the Android app (i.e., *Activity*, *Service*, or *Broadcast Receiver*). The edge set $E$ in a CDG represents the activation relationship between the components. An edge $e = v_{src} \longrightarrow v_{dst}$ indicates that the component $v_{src}$ could activate the start of lifecycle of the component $v_{dst}$.

*5.2.16 Component Behavior Graph.* A **Component Behavior Graph (CBG)** [27] represents the lifetime or control flow logic of the permission-related API functions in a Java or Android program, as well as the functions performed on a particular resource for each component. This is the second half of the CDG, where both of these representations come together to fully describe the Android app. There are four types of nodes, each indicating the type of component at that portion of the graph. The edges connecting the CBG demonstrate the control flow logic between the API functions and sensitive resources.

The CBG $g = (V, E)$ is a directed labeled graph with nodes $v \in V$ and edges $e \in E$. The set of nodes $V$ in a CBG is partitioned into four types: root node $V_{root}$, lifecycle function nodes $V_{life}$, permission-related API function nodes $V_{prapi}$, and sensitive resource nodes $V_f$. A *start node* $v_{root} \in V_{root}$ represents the component itself. A *lifecycle function node* $v_{life} \in V_{life}$ represents the runtime programming logic. Each *permission-related API functions node* $v_{prapi} \in V_{prapi}$ denotes a permission-related API function—for example, Android's API `sendTextMessage()`. A *sensitive resource node* $v_f \in V_f$ indicates sensitive data that is accessed by a component. The edge set $E$ in a CBG represents the control flow logic of the framework API functions and sensitive resources. A CDG edge $e_{cbg} = v_{src} \longrightarrow v_{dst}$ indicates either that if $v_{src}$ and $v_{dst}$ are in the same control-flow block, then $v_{dst}$ is executed right after $v_{src}$ with no executions in between, or if $v_{src}$ and $v_{dst}$ are in two continuous control flow blocks (named $B_{dst}$ and $B_{dst}$, respectively), then $v_{src}$ is the last function node in $B_{src}$ and $v_{dst}$ is the first node in $B_{dst}$.

In Appendix A.2, we provide an example of a CDG and a CBG to demonstrate how they work together.

*5.2.17 Contextual Interprocedural Control Flow Graph.* The **Contextual Interprocedural Control Flow Graph (CICFG)** is an extension of the ICFG, and it describes the complete control flow across all instructions, including context [19]. A *context* defines the information needed for an operation to occur. The CICFG is formally defined as $G = (V, E, \xi)$. The nodes $v \in V$ are basic blocks, and the edges $e \in E$ are either intraprocedural control flows or calling relationships from a node $v_{src}$ to $v_{dst}$. Last, $\xi$ is a set of contexts through which every node $v \in V$ could be reached [19]. The primary difference between the CICFG and the ICFG is that the CICFG gives a more detailed analysis of a program because the context allows to differentiate between different instances or paths that a function may be called. Two examples of a context are *user aware* and *user unaware*, which indicates whether the user is aware of what operations or resources an application or piece of code is using. For example, if an app is using the user's location, in a user-aware context, the user knows that the app is using their location, whereas in a user-unaware context, the user would not know about the app using the user's location [26].

*5.2.18 Contextual API Dependency Graph.* The **Contextual API Dependency Graph (CADG)** is built from a CICFG. Not all of the nodes of the CICFG are security related or invoke a sensitive API. The CADG is an abstraction of the CICFG that only focuses on the security sensitive API invocations [19]. A CADG $g = (V, E, \alpha)$ is a directed labeled graph, with nodes $v \in V$, edges $e \in E$, and labeling function $\alpha$. The set of nodes $V$ in a CADG represents the *basic blocks* of the program. The edge set $E$ in a CADG represents the *data flow* between the basic blocks. A CADG edge $e_{cadg} = v_{src} \longrightarrow v_{dst}$ indicates that $v_{dst}$ uses data that has been defined by the basic block $v_{src}$. The labeling function $\alpha : V \longrightarrow \Sigma$ associates nodes with the labels of corresponding contextual API operations. Each label consists of the API prototype, entry point, and constant parameter [192].

*5.2.19 Contract/Semantic Graph.* A *contract graph* [103] (or a *semantic graph* [108]) is a representation created specifically for vulnerability detection in smart contracts. The set of nodes $V$ in a contract/semantic graph is partitioned into three types: *core node* $V_{core}$, *normal nodes* $V_{norm}$, and *fallback nodes* $V_f$. A core node $v_{core} \in V_{core}$ represents the key invocations and variables that play a crucial role in detecting vulnerabilities. A normal node $v_{norm} \in V_{norm}$ represents invocations and variables that can assist in detecting vulnerabilities, although they do not have the same significance as core nodes. A fallback node $v_f \in V_f$ simulates the fallback function that is incurred on a contract attack.

The edge set $E$ in a contract/semantic graph has three partitions: *control flow* edges $E_{cf} \subseteq (V_{core} \times V_{norm}) \cup (V_{norm} \times V_f) \cup (V_f \times V_f)$, *data flow* edges $E_d \subseteq (V_{core} \times V_{norm}) \cup (V_{norm} \times V_f) \cup (V_f \times V_f)$, and *fallback* edges $E_{fall} \subseteq V_f \times V_f$. A control flow edge $e_{cf} = v_{src} \rightarrow v_{dst}$ indicates that $v_{src}$ can flow to $v_{dst}$ in the next step of the program. A data flow edge $e_d = v_{src} \rightarrow v_{dst}$ indicates that $v_{dst}$ receives data from $v_{src}$. A fallback edge $e_{fall} = v_{src} \rightarrow v_{dst}$ indicates that $v_{dst}$ is the fallback node and $v_{src}$ is the call.value invocation, which is the invocation in a smart contract that can cause a reentrancy vulnerability, if the code is vulnerable [108]. It can also mean that $v_{dst}$ is the function under test, and $v_{src}$ is the fallback node. Generally, this edge indicates interactions with the fallback function [46].

In Appendix A.2, we provide an example of a CADG and a contract/semantic graph for a snippet of source code to demonstrate how the code is abstracted into these graph forms.

*5.2.20 Contextual Permission Dependency Graph.* The **Contextual Permission Dependency Graph (CPDG)** [26] is also built from a CICFG. The CPDG is an abstraction of the CICFG that only focuses on functionality related to Android permissions [26]. A CPDG $g = (V, E, \lambda_p, \xi)$ is a directed labeled graph, with nodes $v \in V$ and edges $e \in E$. Nodes in a CPDG represent the program's *basic blocks* whose functionality pertains to using Android permissions. Edges in a CPDG represent the *data flow* between the basic blocks. A CPDG edge $e = v_{src} \rightarrow v_{dst}$ indicates that there is a path from $v_{src}$ to $v_{dst}$ in the CICFG, and that both nodes are in the same function. $\lambda_p$ is the set of labels representing the concerned permissions. $\xi$ is a set of contexts through which every node is the CPDG could be reached [26].

*5.2.21 Contextual Source and Sink Dependency Graph.* The **Contextual Source and Sink Dependency Graph (CSSDG)** [26] is also built from a CICFG. The CSSDG is an abstraction of the CICFG that considers only the nodes whose functionality is related to using *sources* and *sinks*. [26]. Sources are where sensitive data enters a program, and sinks are where they perform security critical operations. This sensitive data flow could be a point of a vulnerability if the data is not handled properly [96]. Thus, a CSSDG $g = (V, E, \lambda_s, \xi)$is a directed labeled graph, with nodes $v \in V$ and edges $e \in E$. Nodes in a CSSDG represent the basic blocks of the program whose functionality is related to using sources and sinks, whereas the edges represent the data flow between the basic blocks. $\lambda_s$ is the set of labels representing the concerned sources and sinks. $\xi$ is a set of contexts through which every node is the CSSDG could be reached [26].

*5.2.22 Slice Property Graph. Slice property graphs* were proposed by Zheng et al. [111] and aim to preserve the semantics and structural information that is relevant to vulnerabilities. It also aims to eliminate irrelevant information to reduce the complexity of the graphs. The graph uses *SyVCs* (Syntax-based Vulnerability Candidates) as slicing criterion to extract the slice nodes that are relevant to vulnerabilities. Then, edges from the CPG are used as edges between the nodes in the SPG.

*5.2.23 Token Graph. Token graphs* [112] are built from tokens, connecting them via index-focused construction. A token graph $g = (V, E)$ is a directed graph, with nodes $v \in V$ and edges $e \in E$ where $E \subseteq V \times V$. The set of nodes $V$ in a token graph represents individual tokens from the source code. For example, a set of nodes can be $if, x, ==, y$. The edge set $E$ in a token graph defines a co-occurrence relationship between tokens. The co-occurrences describe the relationships between tokens that occur within a fixed-size sliding window [193].

*5.2.24 Propagation Chain.* A *propagation chain* [104] exists when there is a code sequence among a number of specified code snippets. The sequence has direct or indirect data and control dependencies between adjacent code snippets. The propagation chain set $PC(a, b)$ denotes the set of propagation chains between two code snippets a and b. Each program snippet will have propagation chains that affect it and propagation chains that are affected by it. In terms of vulnerability detection, a vulnerable, or defective, propagation chain denotes a code sequence from the vulnerable code to the program's vulnerable output. The set of defect propagation chain, called the *defect propagation chain set*, is denoted as $EPC(d, f)$ and is a subset of the propagation chain set $PC(d, f)$ from a code snippet `d` to the program failure code `f`. Propagation chains can be constructed by data flow or control flow relationships. Zhang et al. [104] use data flow relationships to create the propagation chains for smart contracts. In this instance, the data flow graph is defined as a set of nodes and edges $g = (V, E)$, where the nodes $v \in V$ represent variables in a smart contract and the edge set $E$ denotes the dependency relationships between them. For example, $e = v_{src} \longrightarrow v_{dst}$ denotes that $v_{dst}$ has a data relationship or dependence to $v_{src}$.

### 5.3 Lexical Representations

Lexical representations describe representations that are focused on words and vocabularies. These representations do not show relationships between nodes, as done in graph representations. Lexical representations are also primarily based upon NLP work.

*5.3.1 Tokenizer.* A *tokenizer*, which can also be referred to as a *lexed* representation, takes source code and creates individual tokens for every word or symbol [63]. This is largely based off of existing NLP techniques. While there are a variety of different tokenization algorithms that can be used (e.g., SentencePiece), a majority of the studied papers did not specify what algorithm was used beyond simply stating "the code was tokenized." A few papers, instead, specified the library they used to tokenize the code—for example, Python's tokenizer [117], javalang tokenizer [40], ANTLR [15], js-tokens [120], Clang [137], and phply [86].

For the papers that explicitly stated their tokenization algorithm, we found papers using **Byte Pair Encoding (BPE)** subword tokenization [119, 122], SentencePiece [54], character-level tokenizers [39, 54], WordPiece [129, 138], white space tokenizers [133], and sentence-level tokenizers [87]. BPE subword tokenization is a method that breaks up whole words into smaller parts, in an effort to compress the tokenized data. Frequent words are represented as individual tokens, but infrequent words are split into multiple subword tokens. For example, if the pair of tokens "a" and "b" happen frequently, then they will be combined and become the single token "ab" [194]. SentencePiece is another subword tokenization method and employs lossless tokenization, where all the information needed to reproduce the normalized text is preserved in the output of the encoder (i.e., it treats the input text as a sequence of Unicode characters [195]). WordPiece tokenizes a word using MaxMatch, a process that involves iteratively picking the longest prefix of the remaining text that matches a vocabulary token until the entire word is segmented [196]. In Appendix A.3, we provide an example of BPE subword tokenizer, a standard tokenizer, SentencePiece, and WordPiece for the same source code, to demonstrate the differences between the algorithms.

Tokenizers that use strong embedding algorithms such as `word2vec` [57] are able to capture the semantic meaning of the code. When the vector embeddings are created from the tokenizer, these numbers are largely based off the semantic relationship with another word. In other words, if a word is semantically related to another, their vector representations will be similar [57]. These techniques can be particularly useful when the model needs to learn the semantics of a chunk of code to complete the task at hand. It is also quite simple to tokenize source code, with a complexity of $O(n)$. A built-in function for nearly any language will simply take in a line of text and break it up into tokens based on a provided delimiter. Models such as `word2vec` and `doc2vec` [57, 197] are very developed and are a great way to create word embeddings from a vocabulary. This is a reason this representation is also so popular. However, tokenizers do not capture the structural properties of source code, and this representation thus lacks the ability to understand the syntax of a program.

*5.3.2 iSeVC and sSyVC.* *sSyVCs* (source code and Syntax-based Vulnerability Candidates) are features of code that have some vulnerability syntax characteristics. An example would be for vulnerabilities that are associated with pointers. A sSyVC would be a line of code that contains an asterisk (*) since this symbol is what is used when dealing with pointers [158]. These characteristics are obtained through ASTs. [158]. *iSeVCs* (intermediate code and Semantics-based Vulnerability Candidates) are derived from sSyVC using program slicing. The sSyVCs are the nodes of interest, and the PDG of the program allows one to perform the forward and backward slicing, as described in Section 5.2.7. The resulting set of ordered statements, all containing data or control dependencies between them, are the iSeVCs [158]. iSeVCs contain information regarding data and control dependence, hence their name which relates them to semantics [48].

*5.3.3 Contract Snippet.* A *contract snippet* [47] contains key program statements or lines from a smart contract which could induce a vulnerability. These contract snippets are aimed to be highly expressive such that more pertinent features can be extracted. The contract snippets are all semantically related by control flow dependence, and all highlight a key element (*call.value*) in reentrancy detection (which the paper this representation is proposed in focuses on). Contract snippets can be generated by control flow analysis. Once the contract snippets are created, they are then tokenized and transformed into feature vectors.

### 5.4 Miscellaneous Representations

Miscellaneous representations are those that do not fit into any of the previously described categories.

*5.4.1 Image.* Prior works used the already very developed techniques in image analysis to analyze aspects of code to detect vulnerabilities [87, 92] and malware in Android applications [21]. The core idea behind this method is leveraging visual patterns in software to detect anomalies or similarities. This technique allows researchers to take advantage of the techniques developed for detection of elements in regular images.

*5.4.2 Code Metrics.* *Code metrics* are a quantitative measure that relates certain features to a numerical value, namely the number of times the feature occurs [198]. Code metrics can be defined differently for different tasks. Some common metrics include *lines of code*, *code churn* (i.e., how often code is changed), and more. Prior works also introduce new metrics for a particular purpose, such as SQL injection [156]. Rather than lines of code, or other classical metrics which would not be useful in SQL injection detection, metrics such as number of semicolons, presence of always true conditions, and the number of commands per statement provide more relevant information that would result in better predictions for SQL injection. These metrics can be related to a *risk* factor dictating how much of an impact the metric could have on code to create a security issue [198].

*5.4.3 Code Gadgets.* *Code gadgets* are essentially a method to describe or represent a program slice. They have a number of ordered code statements or code lines that are semantically related to each other by data or control dependency [41]. Code gadgets were created for the exact purpose of vulnerability detection [41], which can explain the reason for its popularity in security-related tasks.

*5.4.4 Opcode Sequences.* An *opcode*, or operation code, specifies the operation to be completed for an instruction [199]. They, in particular, specify the lowest-level operation to be completed such as `PUSH`, `MSTORE`, and `CALLVALUE` [164]. These features can be used to understand on a low level what the code is doing. Opcodes are learned as vectors, and Liao et al. [164] use n-grams and

`word2vec`[57] to learn them as embeddings. Since opcodes already dictate operations in computers, this representation is simple to generate.

*5.4.5 Regular Expression.* A *regular expression* is a sequence of characters that defines a search pattern, often used for string matching or manipulating text within strings based on specified rules. One paper represented source code as regular expressions, which encode information about the API calls in the code. Specifically, the work described by Liu et al. [180] first takes an Android application and creates a CFG from the callbacks. The CFG is transformed to an ICFG, reduced to an API graph, and then an automata to regular expression algorithm is used to generate the regular expressions. This solution is used for multifamily malware classification and addresses the issues of recognizing malware family behavior patterns, code obfuscation, and polymorphic variants that are commonly used by attackers to evade detection. The regular expressions describe the behavior patterns of malware families. While this method can be computationally expensive, as three graphs have to be made before being transformed to a regular expression, it allows to capture the differences between malware families.

*5.4.6 Application Information.* While this representation can be sorted under code metrics, the features extracted by Koli [23] more accurately fall under the name of *application information*. In this representation, an Android application is reverse engineered to extract the original Java files and Android XML. From these files, the API calls made and the permissions used are extracted. Other features such as *is crypto code* or *is database* that specifies certain features of the code that might be associated with malware [23] are also extracted. This representation is a method of feature extraction. Once these features are extracted, they are transformed into vectors of features, and given a label as being benign or malicious. Then, this data can be used in an ML classifier.

*5.4.7 API Calls.* Wang et al. [24] use *API calls* extracted from the source code of an Android application, along with permissions extracted from the Android Manifest file. This paper uses a tool called *DroidAPIMiner* to extract the top 20 API calls that are called by malicious applications. Using these API calls and permissions as features used to train a deep learning model allows [24] to find malware in Android applications. Once again, this representation is a method of feature extraction, and the features are then converted into vectors which are then passed to a model.

**RQ1 Findings:**

- There are 39 representations offering a variety of information about the source code, although a common goal is to capture the semantic and syntactic information in code.
- There are 24 unique graph-based representations, the most popular source code representation category, as it shows the relationships between different nodes (e.g., lines or statements) and how they interact.
- Although there is a larger variety of graph-based representations, 47 papers used tokenizers and 32 papers used ASTs as their representations.
- Seven papers [43–48, 101] propose a representation unique to their application (e.g., a contract graph [46] to find smart contract vulnerabilities). Such task-/language-specific representations are not generalizable to other languages and/or purposes.

## 6 RQ2: Do Certain Tasks Only Use or Mostly Use One Type of Representation?

Figure 2 depicts the relationships between *representations* and *cybersecurity tasks*. *ASTs* and *tokenizers* were the two representations most commonly used for *vulnerability detection*. Given how popular LLMs and NLP have become in recent years, it follows that a tokenizer would be a popular representation, as a tokenizer is what is used for NLP techniques. The vast availability of pre-trained

|  |  |  |
| --- | --- | --- |
| Binary Source Code Matching | Tokenizer | 1 |
| Buffer Overrun Prediction | Tokenizer | 1 |
| Classifying Android Sources & Sinks | Code metrics | 1 |
| | Call graph | 1 |
| Cryptography Misuse | AST | 1 |
| Injection Attack Detection | AST | 1 |
| | Code metrics | 1 |
| | Tokenizer | 1 |
| Malicious Behavior Detection | CDG | 1 |
| | CBG | 1 |
| Malicious Code Classification | AST | 1 |
| | Code metrics | 1 |
| Malicious Code Deobfuscation | AST | 1 |
| | Code metrics | 1 |
| Malicious Code Detection | Tokenizer | 1 |
| Malicious Code Filtering | Tokenizer | 1 |
| Malicious Code Localization | ICFG | 1 |
| | CADG | 1 |
| | CPDG | 1 |
| | CSSDG | 1 |
| | CICFG | 1 |
| Malicious Package Detection | AST | 1 |
| Malware Classification | Regular Expression | 1 |
| | Call graph | 1 |
| | CFG | 1 |
| Vulnerable Code Clone Detection | AST | 1 |
| Malware Detection | Call graph | 3 |
| | ICFG | 3 |
| | CADG | 2 |
| | CPDG | 1 |
| | Image | 1 |
| | Application Information | 1 |
| | API Calls | 1 |
| | Tokenizer | 1 |
| | CSSDG | 1 |
| | CICFG | 1 |
| Malware Prediction | Code metrics | 1 |
| Password Leaks | Tokenizer | 1 |
| Reentrancy Detection | Propagation Chain | 1 |
| | Contract Snippet | 1 |
| | Tokenizer | 1 |
| Security Analysis | AST | 1 |
| Security Patch Identification | Tokenizer | 3 |
| | AST | 1 |
| | Code metrics | 1 |
| Unprotected API Vulnerability Discovery | AST | 1 |
| Vulnerability Analysis | AST | 1 |
| | CFG | 1 |
| | Code metrics | 1 |
| | CPG | 1 |
| | Property Graph | 1 |
| | PDG | 1 |
| | Tokenizer | 1 |
| Vulnerability Classification | AST | 1 |
| | Tokenizer | 1 |
| Vulnerable Commits Detection | Tokenizer | 2 |
| | Code metrics | 1 |
| Vulnerability Detection | Tokenizer | 20 |
| | AST | 19 |
| | CFG | 13 |
| | Code gadgets | 7 |
| | DFG | 7 |
| | PDG | 7 |
| | Code metrics | 4 |
| | CPG | 4 |
| | iSeVC | 3 |
| | AST+ | 2 |
| | Image | 2 |
| | Program slices | 2 |
| | sSyVC | 2 |
| | Contract Graph | 2 |
| | Token Graph | 2 |
| | SCPG | 2 |
| | Call graph | 1 |
| | Semantic Graph | 1 |
| | CDFG | 1 |
| | Program Graph | 1 |
| | VFG | 1 |
| | SDG | 1 |
| | SPG | 1 |
| | CAG | 1 |
| Vulnerability Extrapolation | PDG | 1 |
| Vulnerability Localization | CPG | 1 |
| Vulnerability Prediction | Tokenizer | 7 |
| | Code metrics | 5 |
| | AST | 2 |
| | SDG | 1 |
| Vulnerability Repair | Tokenizer | 3 |
| | AST | 2 |
| Vulnerability Testing | Tokenizer | 2 |
| | Opcode Sequences | 1 |
| | Parse Tree | 1 |

Acronymns
- AST: Abstract Syntax Tree
- CAG: Code Aggregate Graph
- CPG: Code Property Graph
- CBG: Component Behavior Graph
- CDG: Component Dependency Graph
- CADG: Contextual API Dependency Graph
- CICFG: Contextual ICFG
- CPDG: Contextual Permission Dependency Graph
- CSSDG: Contextual Source and Sink Dependency Graph
- CFG: Control Flow Graph
- CDFG: Crucial Data Flow Graph
- DFG: Data Flow Graph
- ICFG: Interprocedural Control Flow Graph
- PDG: Program Dependence Graph
- SCPG: Simplified CPG
- SPG: Slice Property Graph
- SDG: System Dependence Graph
- VFG: Value Flow Graph
- iSeVC: intermediate code and Semantics-based Vulnerability Candidate
- sSyVC: source code and Syntax based Vulnerability Candidate

Legend
tree-based
graph-based
lexical
miscellaneous

Fig. 2. Relationship between representations and tasks

NLP models gives developers the power to easily fine-tune these models on whatever task they desire, without great cost [200]. ASTs are also a relatively cost-effective representation, as compilers use ASTs to represent the structure of source code. This also allows developers to generate this representation with relative ease while also providing crucial structural details about source code that can expose vulnerabilities [175].

Although AST and tokenizers were the most used representations, we also found that vulnerability detection techniques mostly used *graph-based representations*. Graph-based representations were also popular for other cybersecurity tasks, namely malicious code localization, malware detection, vulnerability localization, vulnerability analysis, malware classification, and vulnerability extrapolation. Among the graph-based representations, *CFG* is the most popular one. This could be because it details the program's execution flow, which can be useful in finding whether a vulnerability would occur due to the structure of the program. A developer may choose a CFG over another representation, such as an AST, because it provides more detailed information about the source code. For example, compared to a call graph of a program from the dataset used in Mester and Bodó [35], the CFG has 150,000 vertices/nodes, whereas the call graph has 10,000 nodes. While being more computationally expensive than some other methods (e.g., tokenizer or AST), the information provided by the graph allows for a more robust understanding of code.

*DFG* is another popular representation, particularly for vulnerability detection. Similarly to CFGs, DFGs detail the flow of the program, although they detail the *data flow*. This can also be useful in vulnerability detection because if a harmful data input reaches a security sensitive program point, then a vulnerability will likely occur. Hence, one may choose a DFG over a CFG for detecting a specific type of vulnerability related to data flows. Graph-based representations are useful for these types of tasks because they detail the overall structure and flow of code. To find *sinks*, or points in code where a dangerous function call is invoked, it is helpful to understand the structure because one can pinpoint what function call starts or invokes the vulnerability. Additionally, depending on the type of information shared within the graph, one can better pinpoint areas of code which contains a vulnerability.

*Code gadgets*, *iSeVC*, and *sSyVC* were representations that were specifically made for vulnerability detection techniques. These representations are focused particularly on vulnerability candidates or potentially vulnerable snippets of code. This zoomed-in view allows for a more targeted approach toward vulnerability detection. One may choose this representation for a more lightweight approach to the problem compared to a more resource intensive method such as a CPG.

*ICFGs*, *call graphs*, and *CADGs* are the preferred representations for malware detection. ICFGs describe the complete control flow across a program, and CADGs are derived from a version of ICFGs that include context. Both of these representations provide insights to potential security related invocations, which could allow malware to affect the system. Understanding what calls are made and how instructions flow in a program can give insight to any irregular or unsafe operations done by the code. ICFGs are known to be robust against evasion and obfuscation methods that are used by malware [18]. This could be due to the fact that the ICFG gives the view of the whole program, and malware typically interacts with an entire system beyond just a singular procedure. Rather, malware may call functions outside of a procedure, and the ICFG would give insight to these calls, as well as any unusual or unexpected activity [26].

When choosing a representation for a particular task, it is important to consider the needs and resources of the developer for the application. For example, a *CPG* is a highly robust graph that is complex to generate. If the developer *needs* a highly robust graph for their task and has the *resources* to develop the CPG (for thousands of samples of code), then a CPG would be a good option as a representation. However, if the developer does not have the time or resources to have such a robust graph, but knows that the issue they are trying to detect pertains to the way data flows through a program (e.g., detecting injection attacks), then they might want to choose a DFG, as it would provide the information needed without being overly complex or resource intensive.

**RQ2 Findings:**

- Since vulnerability detection is the most popular task, it is the task that most representations are used for.
- Certain representations (i.e., iSeVC, sSyVC, and code gadgets) were created for the specific task of vulnerability detection and are therefore only used for that task.
- Call graphs, ICFGs, and CADGs are the preferred representation for malware detection.

## 7 RQ3: What Cybersecurity Tasks Are Covered by the ML-Based Techniques?

To better understand the types of tasks that are covered by the techniques discussed in this paper, we sorted the different tasks to fit into the nine disciplines of the **Rational Unified Process (RUP)** cycle [201]. The RUP cycle is a software development process framework that allows software developers to better organize and plan the development process. In this question, we sorted the unique cybersecurity tasks found from our search into the nine main workflows of the RUP cycle: business modeling, requirements, analysis and design, implementation, test, deployment, configuration and change management, project management, and environment. We observed that the cybersecurity tasks only fit into five out of the nine categories—those categories being analysis and design, configuration and change management, environment, implementation, and test. Figure 3 depicts how the different cybersecurity tasks fit into these five disciplines from the RUP cycle.

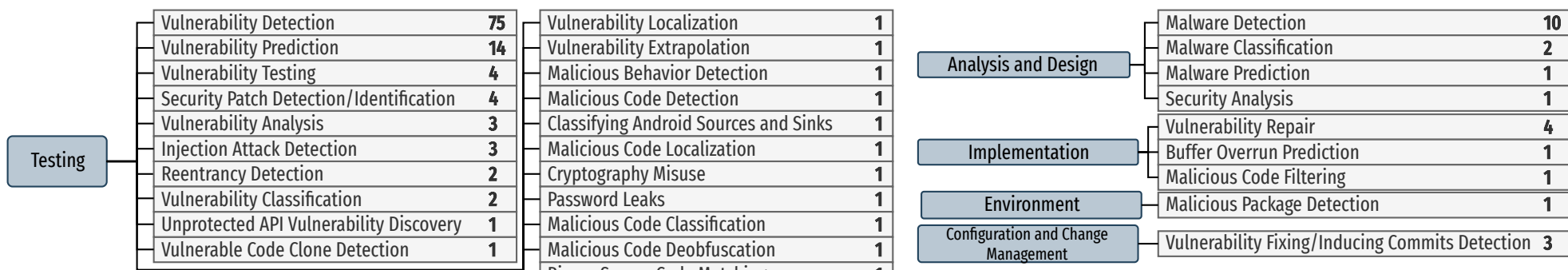

Fig. 3. Cybersecurity Tasks in the RUP cycle

## 7.1 Analysis and Design

Analysis and design involves translating requirements into a formal model of the software, resulting in a system description that guides implementation. We categorize these tasks under analysis and design, as these tasks are analyzing the source code without necessarily implementing a new design. The tasks in this category ensure that, once developers are able to start implementing the system, there are no security issues. *Malware detection*, *prediction*, and *classification* are important tasks, particularly for Android applications [17–26, 35, 150, 180], but also for any system vulnerable to malware. Malware allows attackers to take advantage of security flaws in systems. Understanding types of malware and ransomware, as well as where they occur, allows one to take steps to avoid these issues before implementing a system. *Security analysis* is a task that was used particularly for smart contracts [175], but it can be generalized to any system that has the potential to be compromised.

## 7.2 Implementation

Implementation is the actual coding of the software model. This is when the design from the previous phase is brought to life. *Vulnerability repair* is a task that fixes vulnerabilities in code. This fits into implementation because it is actually implementing the fix to the code while the developer is working [30, 43, 115]. Moreover, there are a few tasks involved with inferring or predicting elements of code, such as *buffer overrun prediction* and *vulnerability prediction*. Buffer overrun is a type of vulnerability, and Choi et al. [128] focus their prediction approach on this one type of vulnerability rather than many others. These prediction techniques allow developers to know where these issues may occur in the code while they are implementing it. We consider these tasks as the implementation phase because vulnerability repair involves implementing a fix to the problem. The prediction tasks, however, are also considered as implementation because it involves being proactive against these vulnerabilities or issues and finding them before they can be exploited.

## 7.3 Testing

Testing involves exercising the software for any flaws or design inconsistencies, helping prevent costly security issues during production. Detection tasks indicate that the vulnerabilities already exist in the implemented system. Thus, the systems are being *tested* for these existing issues so they can be resolved. Many detection techniques fall into this category, such as cryptography misuse, malicious code detection, filtering, classification, deobfuscation, and localization; malicious behavior detection, password leaks, injection attack detection, reentrancy detection, vulnerability analysis, detection, classification, localization, extrapolation, and testing; vulnerable code clone detection; and unprotected API vulnerability discovery. Binary source code matching is also used for tasks like malware detection and vulnerability assessment [202].

## 7.4 Configuration and Change Management

Configuration and change management tracks and maintains a project as it is evolving through time. It ensures that the code created during implementation is still useable and can be reused throughout other portions of the project if needed [201]. Modern software development workflows often use remote repositories to track the code that has been created and its changes. Thus, commits are usually the source of a vulnerability or issue, and the fix to a vulnerability or issue. Zhou et al. [131] and Nguyen-Truong et al. [130] created techniques to assist in the classification of security commits, as well as identifying vulnerability inducing, or vulnerability fixing commits, including security patch detection. Due to the nature of these tasks (e.g., specifically classifying, identifying, or detecting changes relating to security), we categorize these tasks under configuration and change management as they are directly related to how the code is changed and reused throughout a project.

### 7.5 Environment

Environment focuses on the software development environment required for the engineers to develop the system. This includes techniques and processes required by developers [201]. Packages are an essential part of the development process. They provide useful techniques that can greatly simplify the implementation of a system. However, some packages can contain some type of vulnerability or malware that can jeopardize the integrity of the system. Therefore, *malicious package detection* is an important task that can protect systems from such malicious software. We categorize this task under environment because packages are what developers use in their environment as part of their development process. While malicious packages might not necessarily impact their environment directly, it still involves the space in which developers work.

As cryptocurrency becomes a more popular and prevalent topic, research is also starting to focus on creating ML-based techniques to assist in issues relating to cryptocurrencies and smart contracts [46, 47, 79, 101, 103, 108, 164, 166, 175, 176]. All of these papers are interested in vulnerability detection or testing, and security analysis of smart contracts. Two papers [47, 104] focus on a particular vulnerability called *reentrancy attacks*, which is a vulnerability specific to smart contracts. This vulnerability results in an attacker being able to withdraw funds from a smart contract repeatedly and transfer them [47].

While most of the tasks are distinct, some tasks are very closely related to one another. For example, *malicious behavior detection* is looking for suspicious activities and behaviours that might indicate an attack, regardless of if the attack is using malware, whereas *malware detection* is typically looking for specific features or signatures that are known to be associated with malware [17]. *Detection* tasks focus on generally identifying a certain feature (e.g., vulnerability detection focuses on broadly finding a segment of code that is vulnerable), whereas *localization* tasks are focused on pinpointing exactly where, in a larger code base, the feature occurs (e.g., vulnerability localization is focused on pinpointing the specific line in which the vulnerability occurs). Similarly, *prediction* techniques are interested in finding issues before they are apparent in a system (e.g., vulnerability prediction focuses on finding vulnerabilities before they are launched officially into a system, to prevent an attacker having the opportunity to discover it when it is launched), whereas *detection* techniques are interested in identifying issues that already exist in a working system (e.g., vulnerability detection focuses on finding vulnerabilities that are already present in a launched system). *Vulnerability testing*, however, focuses on evaluating a system to determine whether it contains flaws that could be exploited by malicious actors. *Cryptography misuse* describes a detection task, where improper uses of cryptography APIs are discovered. Cryptography APIs tend to be hard to use, so this task alerts developers to when they may have made a mistake in their implementation [33]. *Vulnerability extrapolation* describes a task that aims to find potential vulnerabilities based on existing vulnerabilities. This allows developers to find vulnerabilities that may not exist or be known yet based on past vulnerabilities and their structure. This task in particular focuses on how vulnerabilities can spread or appear in similar contexts based on the vulnerability characteristics [91]. *Unprotected API vulnerability discovery* refers to discovering

Table 3. Languages Supported by Existing Techniques

| Lang. | #Papers | Lang. | #Papers | Lang. | #Papers | Lang. | #Papers | Lang. | #Papers | Lang. | #Papers | Lang. | #Papers |
|---|---|---|---|---|---|---|---|---|---|---|---|---|---|
| **C** | 81 (57.4%) | **JS** | 12 (8.5%) | **Python** | 6 (4.3%) | **C#** | 2 (1.4%) | **Gecko** | 1 (0.7%) | **Powershell** | 1 (0.7%) | **SM** | 1 (0.7%) |
| **C++** | 50 (35.5%) | **Solidity** | 12 (8.5%) | **CSS** | 3 (2.1%) | **SQL** | 2 (1.4%) | **Go** | 1 (0.7%) | **Ruby** | 1 (0.7%) | **XML** | 1 (0.7%) |
| **Java** | 36 (25.5%) | **PHP** | 8 (5.7%) | **Rust** | 3 (2.1%) | **TS** | 2 (1.4%) | **HTML** | 1 (0.7%) | **Smali** | 1 (0.7%) | **XUL** | 1 (0.7%) |

JS = JavaScript; TS = TypeScript; SM = SpiderMonkey.

instances in which a vulnerability can occur when an unprotected API is called or used in a system [174].

**RQ3 Findings:**

- Vulnerability detection by far is the most popular task, with 75 papers focusing on this task.
- Certain papers focus solely on detecting one type of vulnerability, such as buffer overrun prediction, reentrancy detection, and injection attack detection.
- A majority of these tasks fit under the testing category of the RUP cycle, meaning that these techniques are aimed for evaluating the security of already written code before it is deployed.

## 8 RQ4: What Programming Languages Are Predominantly Targeted by the ML-Based Techniques for Cybersecurity Tasks?

Table 3 shows the programming languages that are covered by existing techniques. Two works [35, 125] did not specify the language(s) used. Papers without language or dataset details are labeled as "not specified."

We observed that C is the most popular language; 81 (57.4%) papers developed techniques for C programs. C++ is the second most common language, covered by 50 papers (35.5%), all of which also supported C. The popularity of the C and C++ languages could be attributed to two main factors: the availability of datasets for cybersecurity tasks [203–206], and the higher risk of memory-related vulnerabilities in these languages [207].

We also observed that a number of techniques focus on security tasks for Android applications [12, 12, 17–27, 35, 96, 150, 180]. Although Android applications can be written using `Kotlin` and `Java`, the papers studied in this survey only focused on apps written in `Java`. Additionally, with the increasing popularity of smart contracts, a number of papers developed techniques for `Solidity`, the language in which smart contracts are written.

We also observed that there are only 6 techniques for `Python` and 12 for `JavaScript`, which is surprising given their increasing popularity with developers.[6] If they are used so much in practice, we could expect there to be an associated number of techniques that aim to cover them, particularly for security-related assistance. It is also possible that there are not enough datasets for these languages, and because of this, there are not many techniques for them because researchers are unable to have the necessary data to train and test the models. Nonetheless, this is a gap that should be addressed by future works.

No techniques were language agnostic, meaning that all the systems created were made only for one or potentially a few languages. Making a tool language agnostic is difficult due to the variety of programming paradigms that exist, and because all languages do not follow one particular paradigm. This does mean, however, that researchers should be diligent in developing techniques that support popular and commonly used languages.

[6] https://spectrum.ieee.org/the-top-programming-languages-2023

Table 4. Different Models Used by the Surveyed Papers

| Model Type | Models |
|---|---|
| **Sequence Based** | Transformer [43, 44, 54, 90, 99, 104, 115, 119, 122, 144], Discrete Fourier Transform [51], Convolutional Neural Network [11, 21, 36, 38, 87, 92, 126, 132, 139, 151, 161, 178], Deep Pyramid Convolutional Neural Network [145], Text CNN [39], Temporal CNN [14], Recurrent Neural Network [38, 40, 90, 137, 158], Bidirectional RNN [127, 173], Autoencoders [30], seq2seq [84, 144], BERT [135, 162], codeBERT [129, 138], JavaBERT [138], Hierarchical Attention Network [89, 95, 172], doc2vec [141], word2vec [120, 141], Self attention networks [95], Gated Recurrent Unit [174], Bidirectional Gated Recurrent Unit [86, 160, 171, 178], Online Learning [19], Long Short-Term Memory [11, 117, 123, 134, 135, 152, 159, 163, 165], Bidirectional LSTM [13, 16, 41, 47, 95, 110, 135, 160, 169, 178], Paragraph Vector Distributed Memory [166], Deep Learning Attention-based Convolutional Gated Recurrent Unit [124], Bidirectional RNN for Vulnerability Detection and Locating [48], Passive Aggressive Classifier [18], Extreme Learning Machine [150], Encoder-Decoder [142], GPT [143] |
| **Graph Based** | Graph Neural Network [12, 45, 46, 78, 80, 93, 98, 102, 105, 107, 108, 112–114, 118], Deep Belief Network [24], Graph Convolutional Network [35, 81, 86, 95, 113, 177], Graph Attention Network [97], Feature Attention-Graph Convolutional Network [88], Recurrent Graph Convolutional Network [111], Gated Graph Neural Network [85, 99], Bidirectional Graph Neural Network [37], GraphCodeBERT [101] |
| **Tree Based** | Boosting [155, 156], Extreme Gradient Boosting [125], Light Gradient Boosting Machine [125], adaboost[91], Gradient Boosting Decision Tree [91], Bagging [156], Random Forest [20, 22, 23, 25, 27, 53, 82, 83, 91, 109, 125, 146, 148, 149, 155, 163, 164, 175], coForest [109], Decision Tree [23, 27, 83, 91, 146, 148, 153, 155, 163, 164, 175], code2vec [168], Naive Bayes [17, 23, 27, 53, 83, 125, 148, 153, 163], Tree augmented Naive Bayes [83], Abstract Syntax Tree Neural Network [34], Extra-trees Classifier [36] |
| **Neural Networks** | Multi-Layer Perceptron [22, 79, 82, 91, 94, 107, 114, 136, 152, 170], Neural Network [103, 163, 175, 180], Deep Neural Network [11, 106, 146, 157, 167], Complex Deep Neural Network [146], Abstract Syntax Tree Neural Network [34], Attention Neural Network [100], Neural Memory Network [128], Random Neural Network [132] |
| **Feature Based** | Clustering [91], K-means clustering [141], K-median clustering [179], Naive Bayes [17, 23, 27, 53, 83, 125, 148, 153, 163], Gaussian Naive Bayes, Tree augmented Naive Bayes [83], logistic regression [17, 82, 83, 91, 94, 109, 125, 146, 148, 153, 154, 164], Extreme Machine Learning [150], Nearest Neighbor [20], K-Nearest Neighbor [22, 91, 146, 153, 163, 164], linear regression [146], Support Vector Machine [20, 23, 26, 27, 33, 91, 96, 116, 125, 133, 140, 141, 146, 147, 154–156, 163, 164, 175], C-Support Vector Classification Variant of Support Vector Machine [146], Linear Discriminant Analysis [91, 156, 163], Density-Based Spatial Clustering of Applications with Noise [141] |

**RQ4 Findings:**

— C is the most common language that is targeted by ML-based cybersecurity techniques.

— Despite their popularity, there are not many ML-based techniques for Python and Javascript.

— A large portion of papers are aimed at solving security issues in Android applications. Thus, Java is also a popular language, with 36 techniques geared toward solving these issues in Java.

— Given the increasing popularity of smart contracts, there are a number of techniques (12) that were created for Solidity—the language used to create smart contracts.

## 9 RQ5: What Models Are Commonly Used with Different Representations?

To sort the different types of models used throughout the papers, we took inspiration from Siow et al. [208] and classified them into five categories: *sequence-based* models, *feature-based* models, *tree-based* models, *graph-based* models, and *neural networks*. While there is overlap between our neural network categories and others, our classification is based on the type of inputs that the models accept. For example, a graph neural network may have a neural network architecture; however, it is designed to handle graph-based inputs. Table 4 demonstrates the models used by the surveyed papers and how they fall into each category. By far, a majority of these models are sequence based, particularly CNNs, Transformers, and LSTMs. This is most likely due to their general popularity, but also because these models are very powerful, are able to overcome the vanishing gradient problem, and are able to handle long-term dependencies [209].

However, **Support Vector Machines (SVMs)** overall were the most popular model. This might be because when determining if code has some sort of cybersecurity issue, such as a vulnerability, the most useful thing to learn or focus on are the features of the code. For example, if a model can learn what features make a piece of code vulnerable or not vulnerable, then the model will be able to successfully perform vulnerability detection. SVMs are powerful in learning the features that differentiate classes, or in this case, code. Therefore, SVMs are useful for learning the features of code that would make it vulnerable or malicious versus benign.

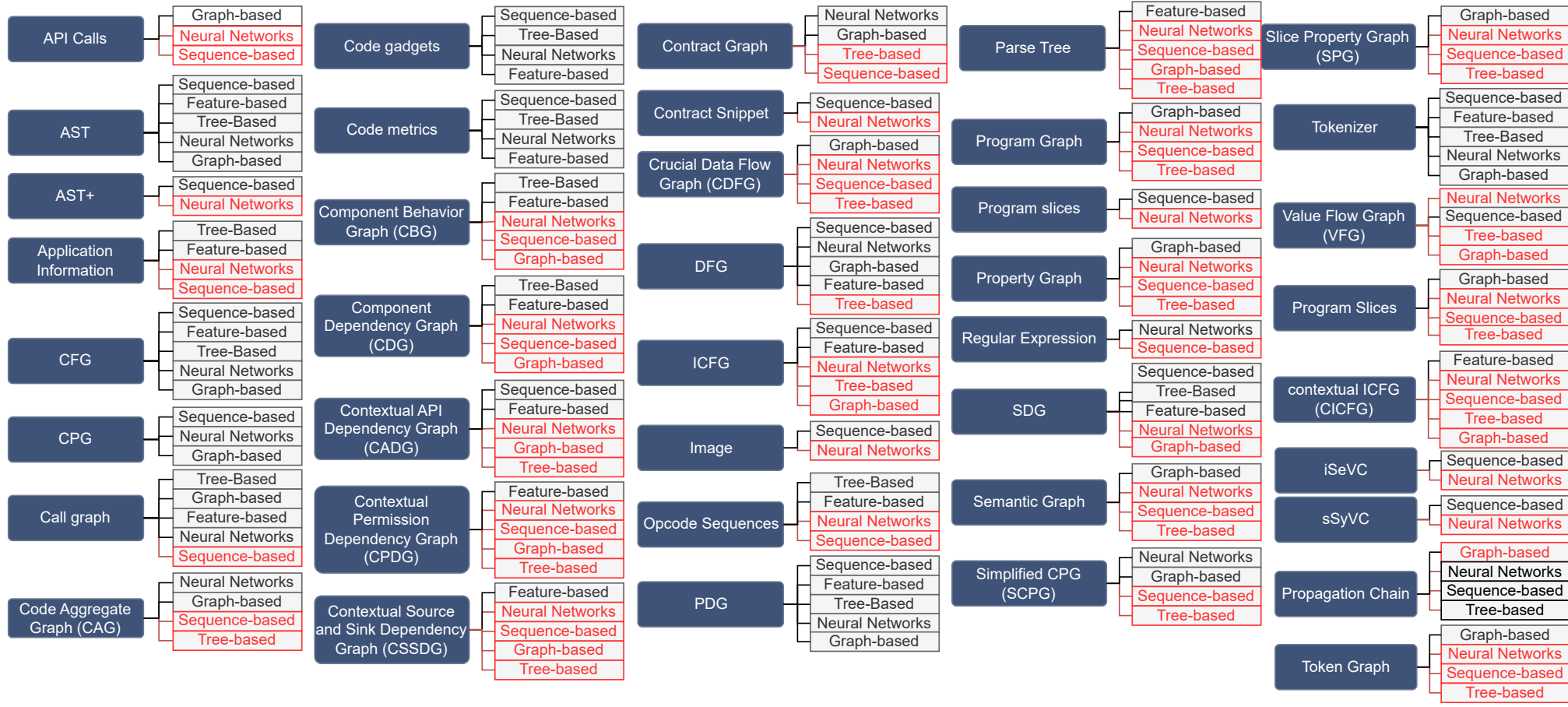

Fig. 4. Relationships between models and representations

Figure 4 shows the types of models that were used for each type of representation. Additionally, the red boxes indicate model types that could potentially be used by these representations, although we did not observe it explicitly in our data. We consider a representation as potentially usable by a model if there are no further transformations needed to get the data into the form the model needs. In other words, this categorization is based on what representations are readily accepted by models. Many papers also used multiple different models to compare the performance of their method. Naturally, there is a constraint on the type of representation and the type of model used. For example, only graph representations can be used with a graph-based model.

**RQ5 Findings:**

- Sequence based is the most common and popular category of models.
- SVMs overall are the most popular model used for different tasks due to their ability to discriminate features that would group code into one group or another.
- The studied papers have a heavier focus on adjusting the models rather than investigating the representation of source code on the whole.

## 10 Threats to Validity

In this section, we will discuss the threats to validity of this survey around construct, internal, and external validity threats as outlined by Runeson and Höst [210].

*Construct validity* refers to how well the operational measures that are used represent the research questions that are being studied. In our study, these measures mainly involve counts of tasks and representations, as well as the relationship between the two, and the relationships between representations and models. Thus, our analysis relies on the accuracy of the reviewers while we were categorizing each of the papers. Additionally, when searching for the papers, we relied on the ability of the search engines we used to return to us all the papers that were related to our search query. To mitigate these issues, we had the reviewers separately analyze each paper, and later discuss and resolve any discrepancies in the analyses. We also created a thorough and broad enough query to ensure the results we received from our search encompassed all the papers related to this study.

*Internal validity* describes how well a study mitigates bias and systematic error such that a causal conclusion can be drawn. Due to the potential threat of incorrectly categorizing a paper, we had two of the authors individually review the papers, and then meet to discuss any discrepancies between the categorizations. Disagreements were resolved through discussion. To assess the reliability of our evaluation, we used Cohen's kappa. Our calculated score is 0.97, meaning that we had a near perfect agreement in our analysis [211].

*External validity* assesses the generalizability of the study. The main threat to this work is that we only focus on the past 10 and a half years (2012–May 2023), so we may have missed papers outside of this range. We also did not include preprints and thus might have missed newer papers. Additionally, our keywords rely on ML, and any papers that do not include these keywords may have been missed. Finally, we only looked at three specific sources: ACM, IEEE, and Springer Link. While there may have been papers outside of these three sources relevant to our work, they were outside of the scope of search for this work.

## 11 Final Considerations

In this section, we discuss our findings, share recommendations for future works, and conclude the work.

### 11.1 Discussion

Aside from our five RQs answered in this SLR, we also identified key findings from our comprehensive analysis related to source code representation:

— *Graph complexity provides more information*: Many papers found that using a graph representation allows to capture more information about the source code, resulting in better performance [37, 41, 44, 78, 81, 86, 89, 93, 98, 105, 106, 110, 113, 114, 135, 158, 161]. Thoroughly capturing the semantic and syntactic relationships and dependencies showed to be crucial for certain tasks, particularly vulnerability detection [41, 44, 78, 81, 86, 89, 93, 98, 105, 106, 110, 113, 114, 135, 158, 161]. Additionally, Sarbakysh and Wang [113] found that using a graph representation allows to reveal hidden connections within the source code, which was found to be useful in vulnerability detection and addressed limitations of traditional methods.

— *Capturing structure/syntactic information is important*: Another common trend is that the structure and syntax of source code is an important aspect that impacts a model's ability to fully learn source code [19, 44, 78, 85, 86, 93, 98, 105, 106, 114, 136, 152, 158, 161, 164, 175]. One study found that in general, graph representations perform better, particularly for vulnerability detection, because the meaningful information relating to structure, syntax, and semantics is preserved through graph representation [98]. In general, one study found that vulnerabilities are related to code structure and patterns [175]. Additionally, other studies found that including some sort of information regarding the structure of the code on top of a representation which does not consider the structure helped improve the overall performance [85, 86].

— *Extracted features impact performance*: While seemingly self explanatory, the features one extracts about the source code will impact the performance of the model because it determines what the model is able to learn. In the case of vulnerability detection, the extracted features may not be associated with some vulnerabilities, thus rendering those features, for lack of a better term, useless to the model [15, 105, 132, 154, 164, 175]. Similarly, certain representations offer more information about the source code than others (i.e., a CPG versus an AST), which enhances the expressiveness of semantics, resulting in more informative extracted features, thus leading the model to perform better [31, 37, 41, 45, 110, 119, 158].

— *Considering code as text does not capture behaviors well*: Many studies agree that treating code as plain text does not provide enough information for a model to learn about the code [43, 44, 81, 89, 105, 119, 132, 135, 136, 158, 164]. Code has more semantic structures that differ from natural language [89], and representing code as plain text causes a loss in this semantic information, such as execution logic and relationships between the code, which is crucial for tasks such as vulnerability detection [44, 81, 105, 132]. One paper found that using BPE subword-level tokenization is more useful than word-level tokenizing [119], as a result of the tendency for code to contain uncommon keywords rather than natural language. Using word-level tokenization, as is done in NLP, results in vector representations that are not particularly meaningful, resulting in poor performance [43, 119].

— *Capturing context improves performance*: Including contextual information helps models better distinguish code features, leading to improved performance [19, 26, 100, 172]. In particular, this contextual information can decrease the number of false positives. In the example of vulnerability detection, added context can assist in precisely distinguishing malicious neighborhoods from benign ones [19].

— *In certain instances, tokenizers can be useful*: While a majority of papers state that tokenizers are not optimal for cybersecurity-related software engineering tasks, two papers point out that there are instances in which tokenizers might be a beneficial representation [15, 122]. Tokenization proves to be useful in making certain syntactic and semantic information easily available to simple neural architectures, which use it for fine tuning on a downstream task [122]. Additionally, one paper found that improving the dictionary by increasing its size and complexity when tokenizing also improved the performance of the method [15]. As previously mentioned, BPE subword-level tokenization is shown to improve performance [43, 119], thus suggesting that while a standard tokenizer (e.g., one used in an NLP model) may not be the most useful representations, there is room to improve tokenization methods such that more pertinent information is passed to the model.

### 11.2 Recommendations

From the 141 papers we have studied, we noticed that rather than attempting to find a new or more comprehensive way to represent source code, prior works are focused on trying different or new models with more advanced architectures so that the model can learn more from the representation, without changing it. Given the improvements in power and capability of ML models in recent years, it is understandable why researchers would take this approach. However, it is important to remember that the way a model learns is impacted greatly by the feature representations, as it can allow the model to learn and isolate critical information that is important for it to successfully complete its task.

Based on the results presented in Sections 5 through 9, we recommend that future works explore different representations for a particular task rather than just fine-tuning the model. Since the model learns features of the source code from the representation, there would be a greater improvement in performance if more attention is paid to the representations. It is also important for researchers to pay attention to the popularity of languages, as there should be an effort to create techniques and tools that address popular and more frequently used languages to ensure that we can avoid as many security risks as possible.

When choosing a representation for a particular task, one might consider the capabilities of different representations. Table 5 highlights these abilities based on key features. "Lightweight"refers

Table 5. Abilities of Different Representations in This Study

| | Structure | Semantics | Data Flow | Statement Flow | Lightweight | Interprocedural |
|---|---|---|---|---|---|---|
| **CPG, PDG, Program Slices, SDG, SPG, CAG** | ✓ | ✓ | ✓ | ✓ | ✗ | ✓ |
| **VFG, Contract/Semantic Graph Program Graph, Propagation Chain** | ✓ | ✓ | ✓ | ✓ | ✗ | ✗ |
| **CADG, Property Graph, CPDG** | ✓ | ✗ | ✓ | ✓ | ✗ | ✓ |
| **CICFG, Call Graph** | ✓ | ✗ | ✗ | ✓ | ✗ | ✓ |
| **SCPG, CSSDG** | ✓ | ✗ | ✓ | ✓ | ✗ | ✗ |
| **AST, Parse Tree, Image** | ✓ | ✗ | ✗ | ✗ | ✓ | ✗ |
| **Tokenizer, Opcode Sequences** | ✗ | ✓ | ✗ | ✗ | ✓ | ✗ |
| **iSeVC, Code Gadgets** | ✓ | ✓ | ✓ | ✓ | ✓ | ✗ |
| **Contract Snippet** | ✗ | ✓ | ✗ | ✓ | ✓ | ✗ |
| **App. Information, API Calls, Code Metrics** | ✗ | ✗ | ✗ | ✗ | ✓ | ✗ |
| **CFG** | ✓ | ✗ | ✗ | ✓ | ✗ | ✗ |
| **DFG** | ✓ | ✓ | ✓ | ✗ | ✗ | ✗ |
| **ICFG** | ✓ | ✗ | ✗ | ✓ | ✗ | ✓ |
| **CDG** | ✓ | ✓ | ✗ | ✗ | ✗ | ✗ |
| **CBG** | ✓ | ✓ | ✗ | ✓ | ✗ | ✗ |
| **CDFG** | ✓ | ✗ | ✓ | ✗ | ✗ | ✗ |
| **Token Graph** | ✗ | ✓ | ✗ | ✗ | ✗ | ✗ |
| **sSyVC** | ✓ | ✓ | ✗ | ✗ | ✓ | ✗ |
| **Regular Expression** | ✓ | ✗ | ✗ | ✓ | ✓ | ✗ |

Representations that have the same abilities are grouped together. BPE is BPE subword tokenization, Op. Seq. is opcode sequences, and App. Information is application information.

to how complex the representation is to create. For example, as discussed earlier, ASTs are simple to generate. Because their computational complexity is relatively low, they can be considered lightweight. Program slices, however, have a lot of overhead and thus are not considered lightweight.

Additionally, one may also want to consider the performance of these representations for particular tasks compared to one another. Figure 5 shows the comparison of the average precision, recall, and F1 scores for certain tasks, as reported by the studied papers. For space considerations, we only demonstrate in this paper the precision, F1, and recall scores, and only for tasks which had multiple representations. The rest of the metrics data in full is available on our GitHub repository for the project.

We can see from Figure 5 that for vulnerability detection, the CPG has the best recall (99.0%), program slices has the best precision (94.8%), and the program graph has the best F1 score (94.5%). Interestingly, all of these representations are graph based, demonstrating the importance of structural information when detecting vulnerabilities. As a CPG is an incredibly robust graph, giving insights to many aspects of a program, it follows that this representation would perform incredibly well. One thing to consider is that because it is such a robust representation, it is complex to generate, which may not be ideal for someone who does not have the resources needed to generate such a graph. Broadly speaking, graph representations show strong results across a variety of tasks, thus speaking to their ability to convey the crucial information relating to the source code. However, in vulnerability prediction and analysis, a tokenized representation did have the best performance compared to other methods (80.1% recall and 96.8% recall, respectively). Code metrics also showed strong results for security patch detection/identification (91.4% F1 score), as well as vulnerability fixing/inducing commits detection (92.0% precision) and vulnerability analysis (96.8% recall). Code metrics are a very lightweight representation, and its performance suggests that the features it presents (e.g., lines of code or code churn) provide enough information about the source code to perform well in tasks relating to security-related code changes and vulnerability analysis. Siow et al. [208] found similar results to us in their empirical study on code representations for code intelligence tasks such as code classification. The paper found that graph-based representations are the best at representing program semantics, thus resulting in the best performance compared to tree-based, feature-based, and sequence-based representations. Additionally, the paper found that different tasks require task-specific semantics to achieve the best performance, but on the whole, a composite graph that represents program semantics comprehensively will produce strong results.

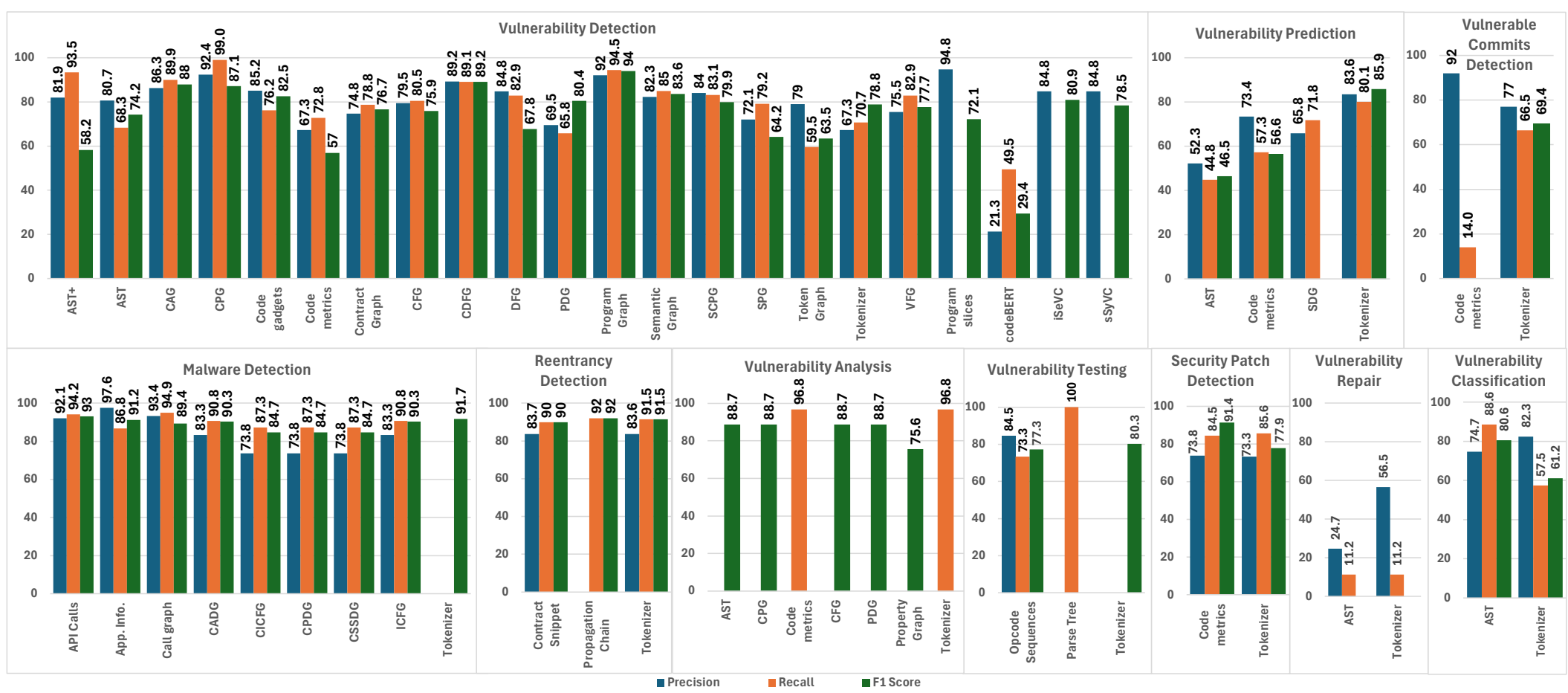

Fig. 5. Aggregated metrics for each representation by task. The abbreviations are as follows: App Info = Application Information; CADG = Contextual API Dependency Graph; CICFG = Contextual ICFG; CPDG = Contextual Permission Dependency Graph; CSSDG = Contextual Source and Sink Dependency Graph; ICFG = Interprocedural Control Flow Graph; AST = Abstract Syntax Tree; CPG = Code Property Graph; CFG = Control Flow Graph; PDG = Program Dependence Graph; CAG = Code Aggregate Graph; CDFG = Crucial Data Flow Graph; DFG = Data Flow Graph; SCPG = Simplified Code Property Graph; SPG = Slice Property Graph; VFG = Value Flow Graph; SDG = System Dependence Graph

### 11.3 Conclusion

In this article, we studied 141 papers to examine the state of the art of source code representations in ML models used for cybersecurity-related tasks. We found that ASTs and a tokenized representation are the most common way to represent source code. We also found that vulnerability detection, malware detection, and vulnerability prediction are the most covered tasks by existing techniques. Additionally, we observed that C was the language covered by the most techniques, followed by C++.

# Appendix

## A Examples of Commonly Used Source Code Representations

### A.1 Examples of Tree-Based Source Code Representations

Figure 6 demonstrates how ASTs and parse trees differ for the same Python code. Notice how the parse tree retained every symbol in the code (e.g., new lines and indentation) while the AST abstracted those away. Since parse trees are more verbose than ASTs, the work [179] that used it had a preprocessing step that removed from the tree nodes that were irrelevant to detect attacks (e.g., specific user ids).

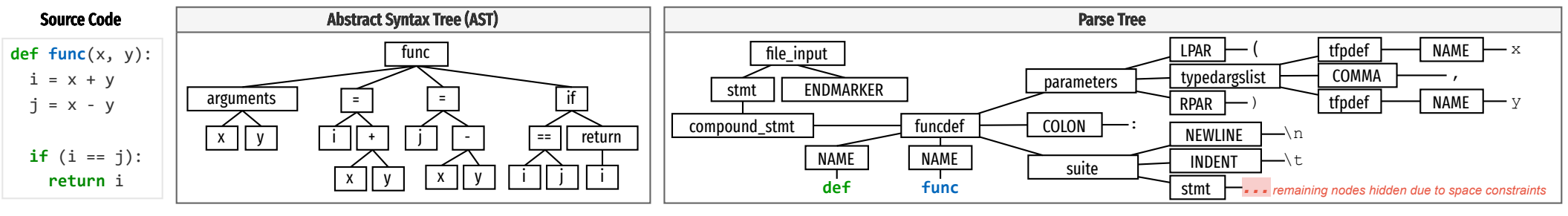

Fig. 6. Examples of tree representations for the same Python source code.

## A.2 Examples of Graph-Based Representations

Figure 7 shows the CFG for the function `func(x,y)`. It has an *entry* and an *exit* basic block, to denote the start and end of the function execution, respectively. The *entry* block is connected to a basic block that has three instructions (i.e., `i=x+y`, `j=x-y`, and `if i==j`). This basic block has two outgoing edges: one represents the flow when the `if` condition evaluates to *true*, and the other denotes the flow when it evaluates to *false*. After executing one of the branches, the execution flows into the *exit* block.

Additionally, Figure 7 shows the ICFG for the entire program, including `func(x,y)` and `main()`. It has two *entry* blocks: one for `main`, where the program starts, and another for `func`. It also has two *exit* blocks which indicate the end of function execution for both `main` and `func`. The *entry* block for `main` is connected to the basic block that calls `func`, which, in turn, is connected to the callee's *entry* block. To capture the flow from the invoked function (i.e., `func(x,y)`) back to its caller (i.e., `main()`), this representation includes an edge from the callee's *exit* block to the caller's *return call site* block. In the same figure, it also shows the DFG for a Python function. There are two outgoing edges `i=x+y` to two statements that use the variable `i`. Similarly, there is an edge from `j=x-y` to `if i==j` because the expression is using the variable `j`.

Figure 7 has an example of a PDG for a Python function (`func(x,y)`), where the dashed and straight lines represent control dependencies and data dependencies, respectively. The PDG starts with an *entry* node for this procedure. It has two *param in* nodes to represent the function's parameters (`x` and `y`). For the statements in lines 2 through 4 to execute, the function's entry must have executed. As such, there is a control dependency edge from the *entry* to the nodes representing these lines of code: `i=x+y`, `j=x-y`, and `if i == j`. Since the statements in lines 2 and 3 define variables that are used in the `if` condition, there is a data dependency edge from these variable assignments statements to the `if` condition. There is also a data dependency edge from the `i=x+y` to the `return` statement, since this `return` node uses the variable `i`. Given that the `return` statement only executes when the `if` condition evaluates to true, there is a control dependency between the `if` node and the `return` node. Finally, Figure 7 shows the call graph for the Python code snippet provided in it.

In Figure 8, the SDG for the entire program is shown. The SDG starts with the *entry* node for `main()`, which is the entrypoint function in the call graph. The first entry node from `main` has a control flow edge to `func`. Two *actual in* nodes have data flow edges, as the data is initialized from `main` and flows to `func`. There are also two *actual in* nodes, which define the parameters entering the procedure `func(x,y)`. The *entry* node in `func` has a control flow edge to the statement nodes which define the variables *i* and *j*, and the two *param in* nodes have a data flow edge to those same two nodes; `i=x+y` and `j=x-y`. These statement nodes then both have a data flow to the predicate node `if i == j`. The *entry* node has a control dependency edge to this node and the *formal in* nodes. Finally, this predicate node has a control flow edge to the `return i` block, along with `i=x+y` node, except this node has a data dependency edge. The dashed line in this figure depicts control dependencies, and the solid line depicts data dependencies. Figure 8 also shows an example of a *backward program slice* over the SDG with the slice criterion $\langle j, line\ 3\rangle$. The nodes that have control or data dependencies to the node of interest (i.e., `j = x - y`) are part of the resulting program slice. In Figure 8, we also see our crucial nodes for a CDFG: any nodes that have to deal with `msg`, `sender`, `balances`, and `_am`. These nodes receive or compute data, and can be responsible for a reentrancy vulnerability.

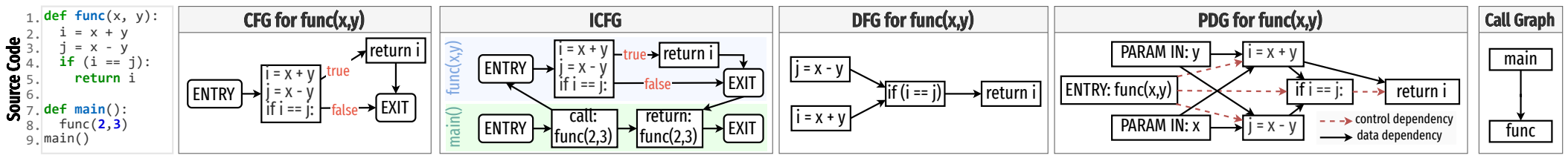

Fig. 7. Examples of graph-based representations (CFG, ICFG, DFG, PDG, and call graph) on the same Python code.

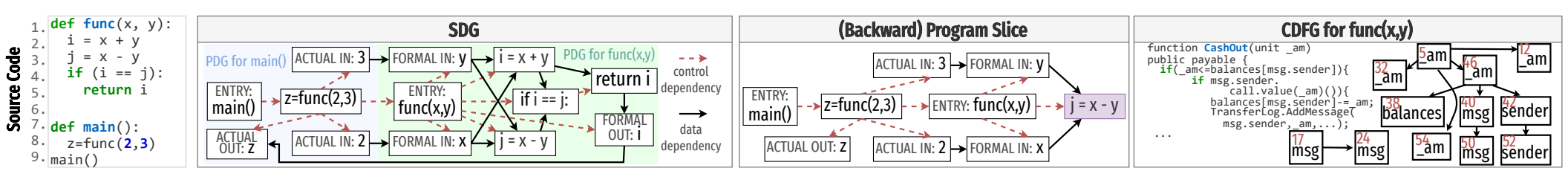

Fig. 8. Examples of graph representations (SDG, Backward Program Slices, and CDFG).

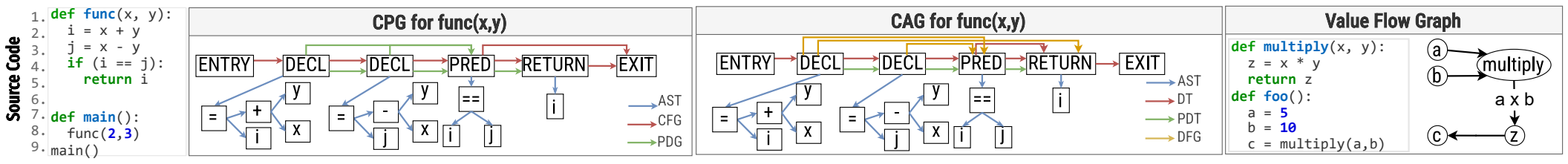

Fig. 9. Examples of graph representations (CPG, CAG, and VFG) on the same code snippet.

Figure 9 demonstrates the CPG for `func()`. We have the *entry* node at the start, demonstrating the entrance into the function. From the entry node, we have a *control flow* edge to the *declaration* node, which then branches to show the *AST* nodes and edges for the line `i=x+y`. We then have another control flow edge from the declaration node to another declaration node. This node branches down to show the AST nodes and edges for the line `j=x-y`. From this node, we have a control flow edge to a *predicate node*. We also have *PDG* edges from the first and second declaration node to this predicate node, as the predicate node depends on the data from the two declaration nodes. The predicate node also has AST nodes and edges to show the line `i==j`. From here, we have a control flow edge to a *return* statement. We also have a PDG edge to this node from the predicate node, as the return statement depends on the result from the predicate edge. We finally have a control flow edge to the *exit* node from the return. The exit node also has a control flow edge from the predicate node because if the predicate evaluates to false, the program terminates. The same figure also demonstrates an example of a CAG for the function `func()`. We start with the *entry* node, which has a *dominator tree* edge to the `DECL` node. The `DECL` node then has *AST* nodes and edges representing the line `i=x+y`. From the `DECL` node, we have a *dominator tree* and *post-dominator tree* edge to another `DECL` node. Once again, the `DECL` node has *AST* nodes and edges representing the line `j=x-y`. Next, we have a dominator tree and post-dominator tree edge to the `PRED` node from the `DECL` node. Moreover, we have a *data dependence* edge from the `DECL` node defining `i` and the `DECL` node defining `j` nodes, as the `if` statement needs the data from both of these nodes to execute. The `PRED` node has a dominator tree and post-dominator tree edge to the `RETURN` node. The `RETURN` node has a *data dependence* edge from the `DECL` node that defines *i*, since the return statement depends on that node's data. Finally, we have a dominator tree edge to the `EXIT` node.

Figure 9 shows the VFG for the code snippet provided. We start with the nodes `a` and `b`. These two nodes have an edge into the `multiply(x,y)` node, as the values from `a` and `b` are passed into this function. We then have an edge to node `z`, which represents the multiplication operation on `a` and `b`. Finally, `z` has an edge connecting it to `c`, which stores the value from the multiplication operation.

Figure 10 shows an example of the CDG which starts at the *component* node. We then have three edges: one for `startActivity()` which leads to a *web page* node, one for `triggerTasks()` which leads to the *background tasks* node, and finally a `sendMessage()` edge which leads to a *message handler* node. Below the CDG, there is an example of the CBG, as a continuation of the larger Web Application Behavior Graph. The *web page* node from the CDG leads to two different nodes in the CBG: `onLoad()` and `onRender()`. Both of these nodes lead to *API* nodes, which can represent any of the aforementioned node types. This pattern continues: the *background tasks* node from the CDG leads to the `beginTask()` node and `complexTask()` node, which then lead to *API* nodes representing the further activities of the APIs in the program. Finally, the *message handler* node from the CDG leads to `onMessageReceive()` node in the CBG, which leads to other API nodes that represent their functionality within the program.

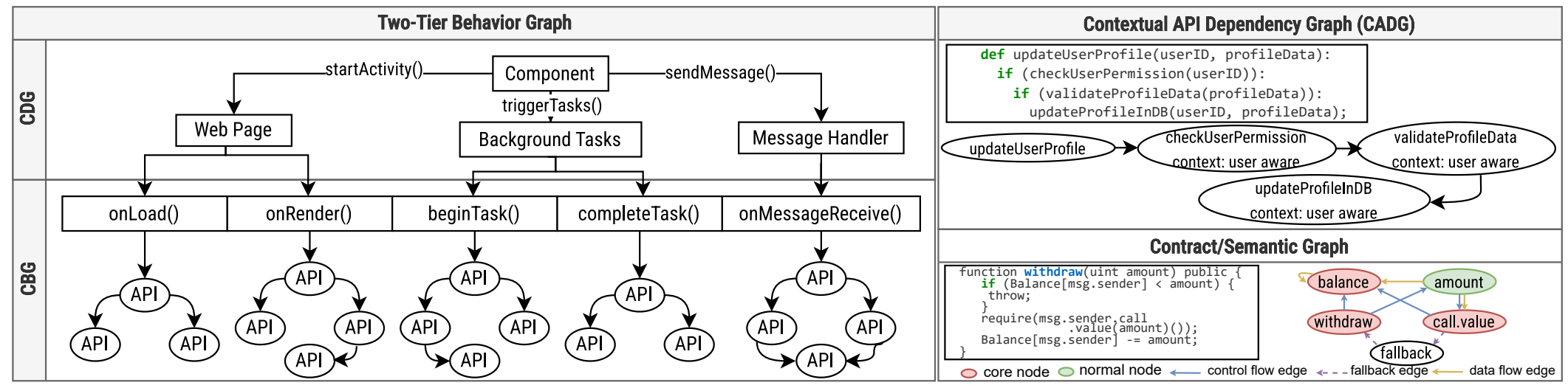

Fig. 10. Examples of CDG, CBG, and CADG.

Figure 10 also gives an example of the CADG for the code snippet shown. We start with the function `updateUserProfile`, and have a control flow edge to the security related node `checkUserPermission` with the context *user aware*. From here, we flow to another security related node, `validateProfileData` with the same context: *user aware*. Finally, we flow to `updateProfileDB` with the same *user aware* context. The same figure also shows a Contract/Semantic graph for the provided code snippet. The `withdraw` and `balance` are core nodes, and the `call.value` variable is a core node because they could be the cause for a vulnerability. The `amount` variable is a normal node, because it does not directly contribute to a security issue. The `withdraw` invocation has control flow edges to the `balance` invocation and the `amount` variable. The `balance` node as a data flow edge to itself since it is accessing data from itself, and receives or depends on data from the variable `amount`. The `call.value` also receives data from `amount`, and thus there is a data flow edge from `amount` to `call.value`. There are control flow edges from `withdraw` to `balance` and `amount` because it invokes these two nodes. After `amount` is called, then `call.value` is invoked, thus resulting in a control flow edge from `amount` to `call.value`. Finally, there is fallback edge from `call.value` to the `fallback` node, and from the `fallback` node back to `withdraw`.

## A.3 Lexical Representations

Figure 11 provides an example of the resulting tokens from different tokenization techniques, including a standard tokenizer, BPE subword tokenization, SentencePiece, and WordPiece.

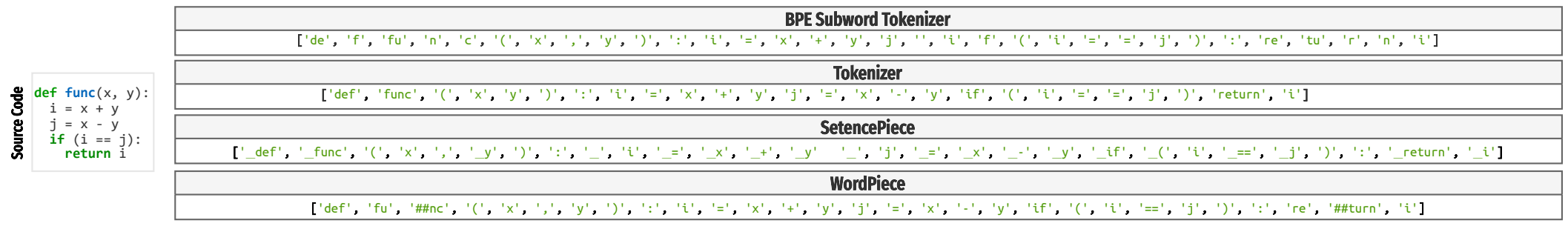

Fig. 11. Examples of different tokenizing algorithms.